\documentclass[final,3p,times,authoryear]{elsarticle}
\usepackage{amssymb}
\usepackage{latexsym}
\usepackage{csquotes}
\usepackage{rotating}
\usepackage{tabularx}
\usepackage{framed,multirow}
\usepackage{tikz}

\usepackage{pgfplots}
\pgfplotsset{compat=1.18}  % Use a version compatible with your TeX distribution

\usepackage{amsmath}

\usepackage{pgfplotstable}
\usepackage{pgfkeys}
\usepackage{url}
\usepackage{xcolor}
\usepackage{svg}
\usepackage{hyperref}
\usepackage[authoryear]{natbib}
\usepackage{subcaption}
\usepackage{amsmath}
\allowdisplaybreaks
\usepackage{algorithm}
\usepackage{algorithmicx}
\usepackage{algpseudocode}
\newcommand{\abssection}[1]{\textit{#1:}}
\usepackage{url}
\usepackage{xcolor}
\definecolor{newcolor}{rgb}{.8,.349,.1}
\setcitestyle{square}

\newcommand\mcs[1][]{\textcolor{pink}{#1}\textcolor{violet}}

\journal{}

\begin{document}

\begin{frontmatter}



\title{Accurate Thyroid Cancer Classification using a Novel Binary Pattern Driven Local Discrete Cosine Transform Descriptor} %% Article title

\author[inst1]{Saurabh Saini}
\ead{phd2101101005@iiti.ac.in}

\author[inst1]{Kapil Ahuja\corref{cor1}}
\cortext[cor1]{Corresponding author}
\ead{kahuja@iiti.ac.in}

%\author[inst2]{Akshat S. Chauhan}
%\ead{akshatschauhan1@gmail.com}

\author[inst3]{Marc C. Steinbach}
\ead{mcs@ifam.uni-hannover.de}

\author[inst3]{Thomas Wick}
\ead{wick@ifam.uni-hannover.de}

\affiliation[inst1]{organization={Math of Data Science \& Simulation (MODSS) Lab, Department of Computer Science \& Engineering},%Department and Organization
            addressline={IIT Indore},
            city={Indore},
            postcode={453552},
            state={Madhya Pradesh},
            country={India}}

%\affiliation[inst2]{organization={Indian Institute of Information Technology},%Department and Organization
%            %addressline={IIT Indore},
%            city={Nagpur},
%            postcode={441108},
%            state={Maharashtra},
%            country={India}}

\affiliation[inst3]{organization={Leibniz Universität Hannover,
    Institut für Angewandte Mathematik},%Department and Organization
            %postcode={},
            addressline={Welfengarten 1},
            city={30167 Hannover},
            %state={Madhya Pradesh},
            country={Germany}}

%% Abstract
\begin{abstract}

\abssection{Background and objectives} Thyroid cancer often manifests as small nodules in the ultrasound images that are difficult to classify manually. The existing Computer-Aided Diagnosis (CAD) systems do not achieve high levels of classification accuracy. Hence, there is need for developing better CAD systems in this domain. 

\abssection{Methods} In this study, we develop a new CAD system for accurate thyroid cancer classification with emphasis on feature extraction. Prior studies have shown that thyroid texture is important for segregating the thyroid ultrasound images into different classes. Based upon our experience with breast cancer classification, we \textit{first} conjecture that the Discrete Cosine Transform (DCT) is the best descriptor for capturing textural features. Thyroid ultrasound images are particularly challenging as the gland is surrounded by multiple complex anatomical structures leading to variations in tissue density. Hence, we \textit{second} conjecture the importance of localization and propose that the Local DCT (LDCT) descriptor captures the textural features best in this context. Another disadvantage of complex anatomy around the thyroid gland is scattering of ultrasound waves resulting in noisy, blurred, and unclear textures. Hence, we \textit{third} conjecture that one image descriptor is not enough to fully capture the textural features and propose the integration of another popular texture capturing descriptor, Improved Local Binary Pattern (ILBP), with LDCT. ILBP is known to be noise-resilient as well. We term our novel descriptor as Binary Pattern Driven Local Discrete Cosine Transform (BPD-LDCT). Final classification is carried out using a non-linear Support Vector Machine (SVM).

\abssection{Results} The proposed CAD system is evaluated on \textit{the only two publicly available} thyroid cancer datasets, namely TDID and AUITD. The evaluation is conducted in two stages (Benign/Malignant and TI-RADS (4)/TI-RADS (5)). For Stage~I classification, our proposed model demonstrates exceptional performance of nearly $\mathbf{100\%}$ on TDID and $\mathbf{97\%}$ on AUITD. In Stage~II classification, the proposed model again attains excellent classification of close to $\mathbf{100\%}$ on TDID and $\mathbf{99\%}$ on AUITD.

\abssection{Conclusion} Our proposed thyroid CAD system delivers superior performance and outperforms all the existing state-of-the-art studies. This will have a big impact in assisting doctors for performing more accurate diagnoses.

\end{abstract}


% \begin{highlights}
% \end{highlights}

%% Keywords

\end{frontmatter}



\section{Introduction}\label{section:intro}
Thyroid nodules (TN) are lumps located in the thyroid region of the neck, which can be either solid or fluid-filled (\cite{cooper2009revised}). They often develop due to excessive production of thyroid hormones like thyroid stimulating hormone (TSH), T3, and T4. Although most thyroid nodules are benign and generally harmless, some have the potential to become cancerous (\cite{srivastava2024deep}). Over the past thirty years, the incidence of thyroid cancer has surged by about $240\%$, making it one of the fastest-growing cancers in terms of new cases (\cite{song2018multitask}). This substantial rise emphasizes the importance of monitoring thyroid health and raising awareness about potential risks.

Ultrasonography (USG) has become the primary and most popular method for diagnosing thyroid nodules. It is frequently employed to assist in fine-needle aspiration biopsy (FNAB) and subsequent treatments. Although other imaging techniques such as computer tomography (CT) scan and magnetic resonance imaging (MRI) are available, USG remains highly recommended for detecting thyroid nodules due to its lack of radiation, cost-effectiveness, and real-time imaging capabilities (\cite{srivastava2022hybrid}).

Recent guidelines have been developed to help radiologists assess thyroid nodules using ultrasound characteristics (\cite{kwak2011thyroid}). However, diagnosing thyroid nodules with USG can be challenging due to issues like echo disturbances and speckle noise, which means that an accurate diagnosis heavily depends on the expertise and precision of experienced radiologists. Inexperienced practitioners are more likely to misinterpret these ultrasound characteristics, leading to higher rates of misdiagnosis. This can result in unnecessary biopsies and surgeries, increasing patient stress and anxiety, as well as medical costs. To address this, there is a pressing need for an advanced Computer-Aided Diagnosis (CAD) system for thyroid nodules that acts as an expert.

In a CAD system, feature extraction represents the most critical component and can be carried out in two ways. First, by using deep learning descriptors (\cite{song2018multitask, srivastava2024deep, saini2024improved}), which have shown superior accuracy and robustness but come with several limitations. These include the need for large amounts of labeled training data, reliance on high-performance hardware, high computational complexity, and not being inherently explainable, which limits their adaptability in real-world clinical scenarios (\cite{wang2025robust, karanwal2024robust}). Second, by employing handcrafted feature extraction techniques (\cite{ALI201839, MAZO20171, ESTEBAN2019303}), which are typically designed based on domain expertise. Their major advantages are that they can be effective even when labeled training data is limited, require less sophisticated hardware, are computationally efficient, and explainable, making them more practical in real-world medical settings. However, their main limitation is that they are generally less accurate and robust. As evident, the handcrafted descriptors have more advantages than deep learning-based approaches, and if we can make them more accurate and robust, then they will make the best CAD systems.

 In the past, several handcrafted descriptors, such as Wavelet Multi-sub-bands Co-occurrence Matrix (WMCM), Histogram of Oriented Gradients (HOG), Local Binary Pattern (LBP), and Gray-Level Co-occurrence Matrix (GLCM), have been used to capture different types of features, including edges, texture, shape, and structure. These descriptors are typically combined with classification algorithms such as Support Vector Machines (SVM), K-Nearest Neighbors (KNN), and Random Forests (RF) (\cite{dandan2018texture, colakoglu2019diagnostic, wu2019classification}). However, as expected, these methods have not achieved high performance (due to the inherent nature of the handcrafted descriptors), which we address in this work.

It is well-known that texture is important for classifying thyroid nodules (\cite{tasnimi2023diagnosis, poudel2019thyroid}), however, we conjuncture that it is the most critical feature for it. In one of our recent work on breast cancer, where texture also played a key role, the Discrete Cosine Transform (DCT) descriptor outperformed all other handcrafted descriptors (\cite{shastri2018density}). Drawing intuition from that study, our \textbf{first design decision} is to focus on the DCT descriptor.

Moreover, thyroid imaging can be difficult because of the complex anatomy of the neck. Structures such as the recurrent laryngeal nerve, trachea, and esophagus lie close together, causing variations in tissue density and resulting in overlapping regions. Hence, we conjuncture that designing a descriptor in a local context can better capture texture. Based upon this intuition, our \textbf{second design decision} is to employ a local variant of DCT, namely the Local Discrete Cosine Transform (LDCT) descriptor.

The presence of multiple complex anatomical structures around the thyroid gland also scatters the ultrasound waves, leading to noisy and blurry images with unclear and distorted textures (\cite{gervasio2010ultrasound}). Therefore, we conjuncture that a single descriptor is insufficient to capture the full range of textural information. Hence, we integrate the Improved Local Binary Pattern (ILBP) descriptor, which is another popular available texture-exploiting descriptor, with LDCT. ILBP is known to be noise resistant as well. This is our \textbf{third design decision}, resulting in our novel Binary Pattern Driven Local Discrete Cosine Transform (BPD-LDCT) descriptor. As demonstrated in the results section discussed later in the manuscript, all these design choices lead to excellent results.

The proposed CAD system has four components, namely preprocessing, feature extraction, data balancing, and classifier.

$\mathrm{\textit{(I)}}$ \textit{Preprocessing.} This is a standard component in a CAD system. We first segment the thyroid nodule image using image binarization to extract the relevant region. Next, we perform normalization, as these images are captured under different lighting conditions. Finally, we apply Two-stage Contrast Limited Adaptive Histogram Equalization (TS-CLAHE) to enhance the texture of the thyroid nodule image.

$\mathrm{(\textit{II})}$ \textit{Feature Extraction.} In this component, the thyroid nodule image is divided into non-overlapping cells, and LDCT is applied to each cell individually. It transforms each spatial cell into the frequency domain, producing a set of coefficients that are subsequently used for texture analysis. Next, we apply ILBP on the frequency coefficients of each cell obtained from LDCT, and subsequently generate a binary pattern with its corresponding decimal representation, resulting in our \textit{novel} BPD-LDCT descriptor.

$\mathrm{(\textit{III})}$ \textit{Data balancing.} Generally available datasets in this domain are imbalanced, hence, we address this issue of class imbalance by the Synthesis Minority Oversampling approach (SMOTE) (\cite{liu2024feature, chawla2002smote}). It balances the dataset by generating synthetic minority samples through interpolation between neighboring minority instances, rather than simply duplicating existing samples, which enhances classifier learning.

$\mathrm{(\textit{IV})}$ \textit{Classifier.} For classification, we employ a non-linear SVM to capture complex, non-linear relationships between the extracted features and class labels. This choice is motivated by its ability to efficiently handle the high-dimensional feature space generated by our BPD-LDCT descriptor, while also maintaining robustness against overfitting in small-sample scenarios.

We evaluate the performance of our proposed CAD system \mcs{on} \textit{the only two publicly available} datasets, namely the Thyroid Digital Image Dataset (TDID) and the Algerian Ultrasound Images Thyroid Dataset (AUITD) across two stages. In Stage $\mathrm{I}$, we classify thyroid nodules as benign or malignant, while in Stage $\mathrm{II}$ malignant nodules are further categorized based upon their TI-RADS score, a rating system used to estimate cancer risk. For Stage $\mathrm{I}$, we achieve an excellent average performance of around $\textbf{100\%}$ and $\textbf{97\%}$ on TDID and AUTID, respectively. For Stage $\mathrm{II}$, we again achieve a strong average classification of close to $\textbf{100\%}$ and $\textbf{99\%}$ on TDID and AUTID, respectively. Overall, our approach significantly outperforms existing state-of-the-art methods.

The remainder of this manuscript has four more sections. In Section \ref{section: Literature_2}, we review the existing literature. In Section \ref{section: Methodology}, we present our proposed model. In Section \ref{section: result}, we give the details about the datasets and the experimental results. Lastly, in Section \ref{section:conclusion}, we present a conclusion and discuss the future directions for this work.

\section{Literature Review} \label{section: Literature_2}
An overview of published results for Stage $\mathrm{I}$ classification (benign or malignant) is given in Table \ref{tab:Literature_survey}. No one has attempted Stage $\mathrm{II}$ (based on TI-RADS) classification, and hence, there is nothing to report here. Table \ref{tab:Literature_survey} is divided on the basis of the feature extraction approaches, whether it is deep learning based or handcrafted. We also give the following details: the specific techniques used for feature extraction and further classifying them; the datasets along with the number of images tested upon; the type of data balancing performed; and the accuracy of the resulting classification. All the data in this table is self-explanatory, however, for each work, we summarize the details of the feature extraction technique.

In the deep learning category, \cite{song2018multitask} proposed a Multi-task Cascade Pyramid Convolutional Neural Network (MC-CNN). This model captured the low-level features like edges and texture using the initial convolution layers and high-level features like shape and structure using the spatial pyramid convolutional layers. \cite{nguyen2019artificial} used ResNet18, ResNet34, and ResNet50, to extract the features from the spatial domain and Fast Fourier Transform (FFT) to extract the features from the frequency domain. Again, \cite{nguyen2020ultrasound} used ResNet50 and Inception models to capture the textural and structural features. \cite{hang2021thyroid} used ResNet18 and several handcrafted methods such as Speeded-Up Robust Feature (SURF), Scale-Invariant Feature Transform (SIFT), HOG, and LBP to capture the textural and structural features. \cite{srivastava2024deep} used AlexNet and VGG16 to extract low-level and high-level features.

In the handcrafted category, \cite{dandan2018texture} captured the textural features using the proposed Wavelet Multi-sub-bands Co-occurrence Matrix (WMCM). This enhanced the visibility of subtle changes despite the speckle noise common in ultrasound images. \cite{colakoglu2019diagnostic} used Histogram of Oriented Gradients (HOG) to capture edge orientations and Local Binary Pattern (LBP) to capture the textural patterns. Finally, in this category, \cite{wu2019classification} also used HOG again to capture edge orientations and Gray Level Co-occurrence Matrix (GLCM) to capture the spatial relationships between pixel intensities.

\renewcommand{\arraystretch}{1.3}
\begin{table*}[!h]
	\centering
	\resizebox{\textwidth}{!}{
		\begin{tabular}{c|l|l|c|l|c|l}
			\hline
			Method& Ref. & Extraction & Classifier& Datasets& Data  & Accuracy\\
			&&methods&&&balancing& \\\hline

			\multirow{22}{*}{Deep Learning}&\cite{song2018multitask}&MC-CNN&End-to-End& Private& Cropping & $92.10\%$\\
			&&&&$6228$ images& &\\
			&&&&TDID&&\\
			&&&&$152$ images& & \\
			&&&&&&\\
			&\cite{nguyen2019artificial}&ResNet50,&End-to-End &TDID& No& $ 90.88\%$ \\
			&&ResNet18,&&$450$ images& &   \\
			&&ResNet34,&&&&  \\
			&&FFT&&&&\\
			&&&&&& \\

			&\cite{nguyen2020ultrasound}  &ResNet50,&End-to-End&TDID& Weighted Binary  & $92.05\%$\\
			&&Inception&&$450$ images&Cross Entropy& \\

			&&&&&&\\

			&\cite{hang2021thyroid} &Res-GAN,&RF&TDID&Rotation,&$95.00\%$\\
			&&SURF, LBP, &&$428$ images&Flipping& \\
			&&SIFT, HOG&&&& \\
			&&&&&& \\

			&\cite{srivastava2024deep} &AlexNet, &End-to-End&Private  & GAN &$96.00\%$\\
			&&VGG-16&&$654$ images& & \\
			&&&&TDID&& \\
			&&&&$295$ images&&\\
			&&&&&&  \\

			\hline
			\multirow{8}{*}{Handcrafted}&\cite{dandan2018texture}&  WMCM &SVM,& Private &No&  $87.00\%$ \\
			&&&KNN&$288$ images& &\\
			&&&&&&\\

			&\cite{colakoglu2019diagnostic} &HOG, &RF&Private&No &$86.80\%$\\
			&&GLCM& & & & \\

			&&&&&&\\
			&\cite{wu2019classification} &HOG, LBP&SVM &Private &No &$88.59\%$  \\
			&& &&$4574$ images&&   \\

			\hline

	\end{tabular}}
	\caption{Summary of previous state-of-the-art studies available in deep learning and  handcrafted category for benign or malignant thyroid nodule cancer classification.}
	\label{tab:Literature_survey}
\end{table*}

\textit{ In summary, the literature suggests that deep learning descriptors deliver strong performance as compared to handcrafted descriptors. Since the handcrafted descriptors have the advantages of requiring less amount of data, needing less sophisticated hardware, and being computationally efficient and inherently explainable, we focus on them. In this work, we propose a novel handcrafted descriptor that beats the performance of the best deep learning descriptor. }

\section{Proposed thyroid nodule classification system}\label{section: Methodology}
\begin{figure*}[!h]
	\centerline{\includegraphics[width=\textwidth]{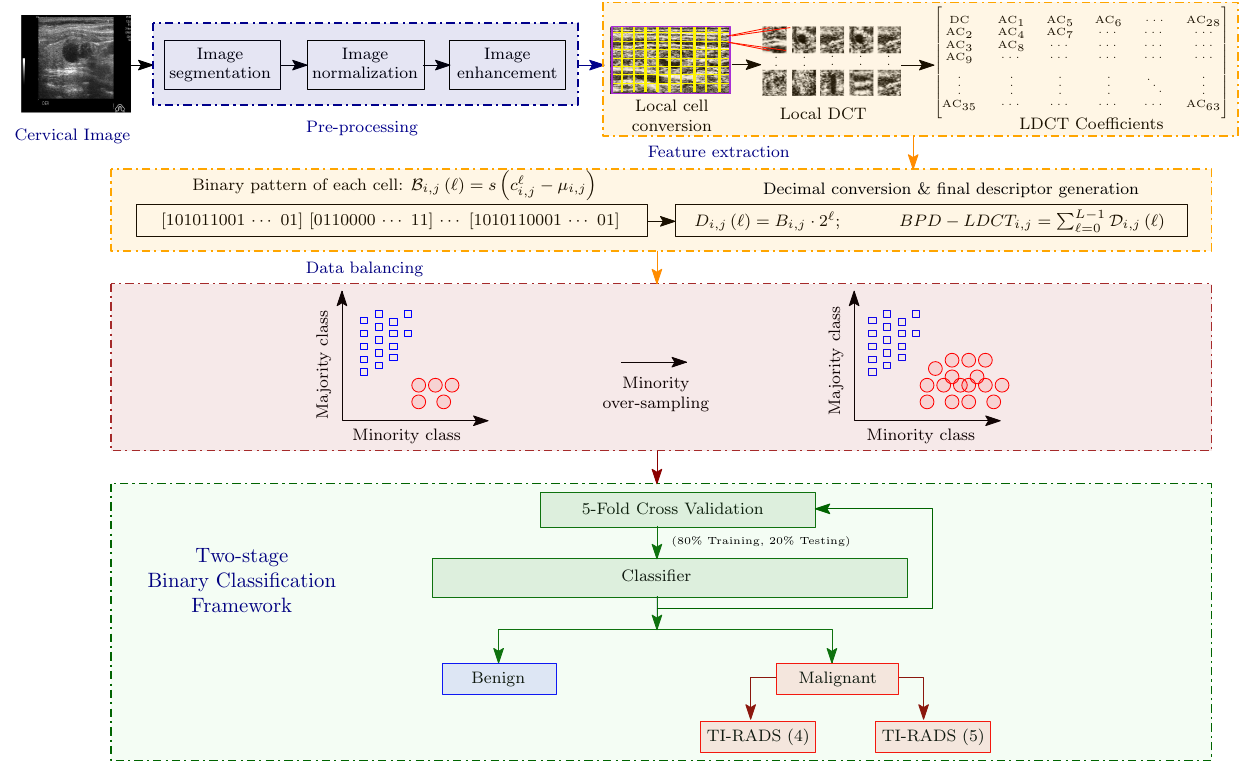}}
	\caption{Proposed framework of our two-stage thyroid cancer classification model.}
	\label{fig: proposed model}
\end{figure*}

In this section, we present our two-stage thyroid nodule classification model. Figure \ref{fig: proposed model} shows the complete flowchart of our proposed model, which consists of four key components: preprocessing, feature extraction, data balancing, and classification. In preprocessing, we extract the region of interest from the thyroid nodule images and perform image normalization and enhancement. In the feature extraction, we propose our novel BPD-LDCT descriptor to capture the meaningful textural features from the complex thyroid nodule images. For oversampling, we apply the SMOTE that balances the number of samples of each class. Finally, we use a non-linear SVM with a Radial Basis Function (RBF) kernel to classify the represented features into the corresponding classes. A detailed description of each component is given in the subsequent subsections.

\subsection{Pre-processing of thyroid nodule ultrasound images} \label{subsection: preprocessing}

\begin{figure*}[!h]
	\centerline{\includegraphics[width=0.55\columnwidth]{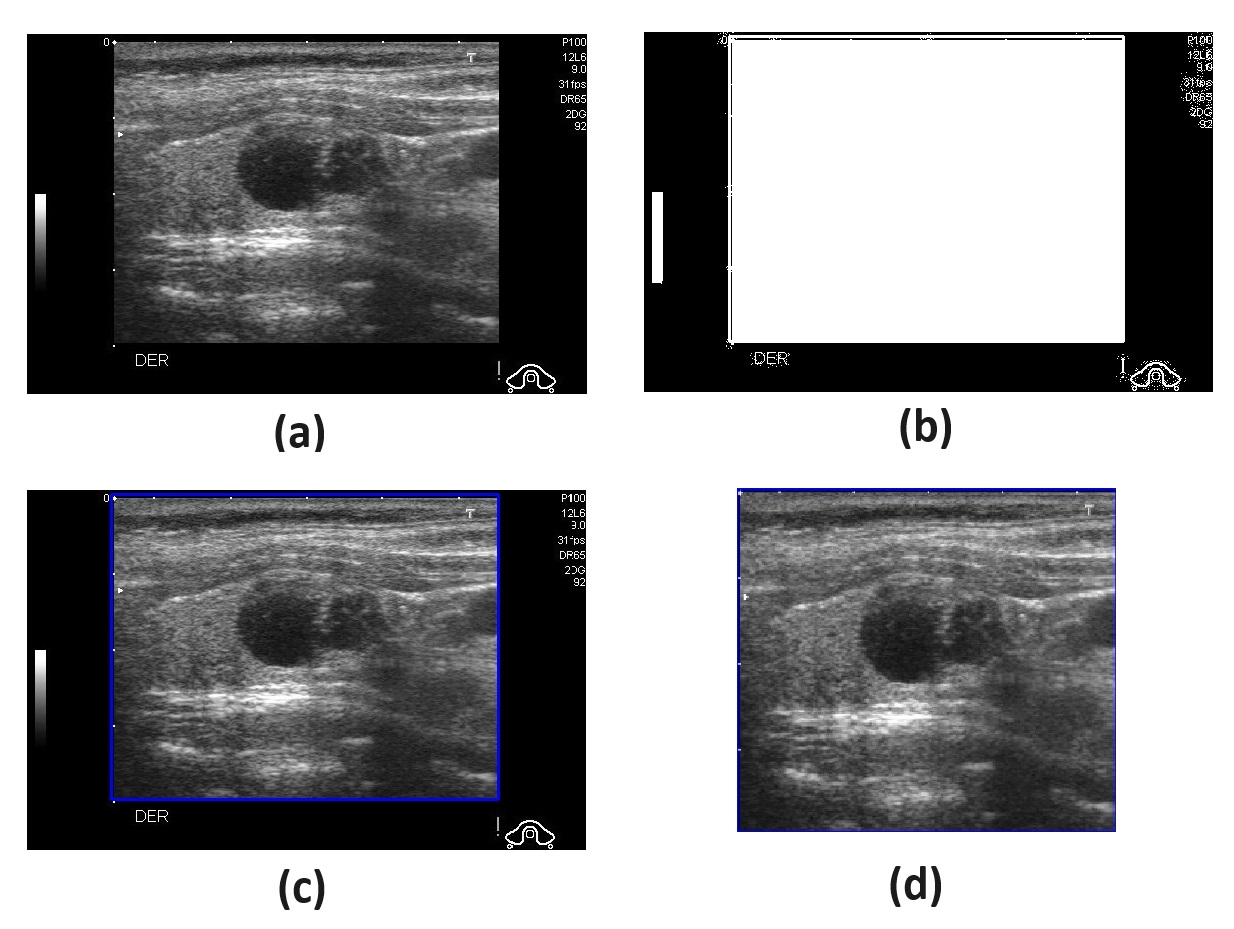}}
	\caption{\scriptsize Result of thyroid nodule region detection algorithm used in our work: (a) Input ultrasound thyroid image (b) Binarized thyroid image (c) Thyroid region detection (d) Final result of detected thyroid region. }
	\label{fig: binarize}
\end{figure*}

Each thyroid sample consists of two parts; the inner thyroid region, which is our region of interest, and the background, containing diagnostic information for the physician or noise that can deteriorate the quality of the feature vector, see Figure \ref{fig: binarize}(a) as an example. In preprocessing we extract this region of interest. For this, we use image binarization (\cite{otsu1979threshold, nguyen2019artificial,nguyen2020ultrasound}), which converts a grayscale image into a binary image by classifying pixels as either “foreground” (region of interest) or “background” based on a threshold. Pixels brighter than the threshold turn white, and darker pixels turn black, resulting in a binary image with two colors as depicted in Figure \ref{fig: binarize}(b). Here, the largest rectangular white area corresponds to the thyroid nodule as in Figure \ref{fig: binarize}(c). This white region is then extracted as the thyroid nodule patch, as illustrated in Figure \ref{fig: binarize}(d).

The inconsistency in lighting conditions during the ultrasound image-capturing process leads to variations in the grey-level intensity of different thyroid patches. Next, we address this issue via normalization. This method standardizes the intensity of each pixel within a patch by scaling the values to a uniform range between 0 and 1, ensuring consistency across the patches (\cite{jain2005score}). It is formulated as follows:
\begin{equation}
I^{*}(x,y)  = \frac{I(x,y) - I_{min}}{I_{max} -  I_{min}}, \nonumber
\end{equation}
where $I^{*}(x,y)$ represents the normalized intensity value of the pixel at location $(x,y)$, $I(x, y)$ represents the original intensity value of the pixel at location $(x,y)$, and $I_{max}$ and $I_{min}$ represent the maximum and minimum intensity values across all the pixels.

\begin{figure*}[!h]
	\centerline{\includegraphics[width=0.80\columnwidth]{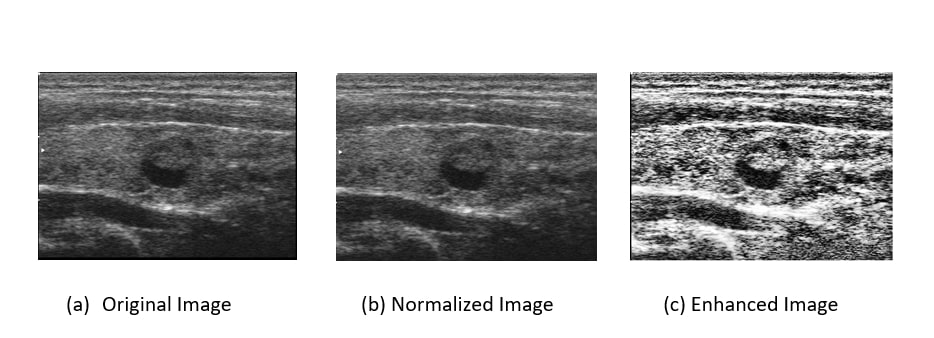}}
	\caption{\scriptsize Normalized and Enhanced thyroid nodule patch. }
	\label{fig: Normalize and enhanced}
\end{figure*}

Since the thyroid nodule patches have low contrast, leading to difficulty in capturing the texture (\cite{morin2015motion}), we enhance it. A basic method for contrast enhancement is Histogram Equalization (HE), which increases contrast in the bright areas and reduces it in the dark areas. However, for images where multiple objects of interest are present, like in thyroid nodules, HE may not provide a sufficient enhancement (\cite{morin2015motion}). To overcome this limitation, more advanced techniques like Adaptive Histogram Equalization (AHE), Contrast Limited Adaptive Histogram Equalization (CLAHE), and Two-Stage Adaptive Histogram Equalization (TSAHE) are often used (\cite{anand2015mammogram,panetta2011nonlinear}). Two stages of CLAHE have been found to be particularly effective for enhancing small tissue structures, as seen in mammogram patches for breast cancer (\cite{shastri2018density}). Hence, we apply a two-stage CLAHE process to the thyroid patch. First, CLAHE is applied to $8 \times 8$-sized blocks, followed by a second application on $4 \times 4$-sized blocks. Figure \ref{fig: Normalize and enhanced} illustrates the original, normalized, and enhanced images, clearly showing that the texture of the thyroid patch is significantly improved after applying the two-stage CLAHE.

\subsection{Feature extraction}\label{subsection: extraction}
As discussed in Section~\ref{section:intro}, texture is the most important feature in thyroid cancer, and we choose the Discrete Cosine Transform (DCT) based upon its proven effectiveness in a previous breast cancer study (\cite{shastri2018density}). Hence, initially we provide the working of the DCT descriptor in Section~\ref{subsubsec: DCT}. Due to the complex anatomy of the thyroid region, where multiple neighboring structures and varying tissue densities introduce significant heterogeneity, local analysis is important. Hence, next we present the working of the Local Discrete Cosine Transform (LDCT) descriptor in Section~\ref{subsection:LDCT}. Furthermore, again due to the presence of several anatomical structures around the thyroid gland, ultrasound imaging leads to noisy and blurry texture in the resulting images. Since in this context it is challenging for a single descriptor to capture all the textural features, we integrate Improved Local Binary Pattern (ILBP), another popular texture-exploiting descriptor with LDCT. ILBP is also known to be noise resistant, making it ideal for such an integration. Finally, we present the working of ILBP in Section~\ref{subsubsection:ILBP}, and details of our novel Binary Pattern Driven Local Discrete Cosine Transform (BPD-LDCT) descriptor in Section~\ref{subsubsection:BPDLDCT}.

\subsubsection{Discrete Cosine Transform (DCT) descriptor} \label{subsubsec: DCT}
Given a thyroid nodule patch $\mathtt{I}$ with pixel intensity \( {I}(x, y) \) at position \( (x, y) \), where \( x = 0, 1, \ldots, N-1 \) and \( y = 0, 1, \ldots, N-1 \), the frequency representation of the patch obtained by applying the DCT is defined as follows (\cite{shastri2018density}):

\begin{align}
	F(u, v) = \alpha(u)\alpha(v) \sum_{x=0}^{N-1} \sum_{y=0}^{N-1} {I}(x, y)
	\cos\left[\frac{(2x+1)u\pi}{2N}\right] \cos\left[\frac{(2y+1)v\pi}{2N}\right], \notag \\
	\label{eq:global_dct}
\end{align}
\noindent where \( u \) represents the horizontal frequency index with \( u = 0, 1, \ldots, N-1 \) and \( v \) represents the vertical frequency index with \( v = 0, 1, \ldots, N-1 \). The \( \alpha(u) \) and \( \alpha(v) \) are the normalization terms and are defined as follows:

\begin{equation}
	\alpha(0) = \frac{1}{\sqrt{N}}, \qquad
	\alpha(k) = \sqrt{\frac{2}{N}} \quad \text{for } k > 0.
	\label{eq:normalization}
\end{equation}

The DCT coefficients are generally grouped into three frequency bands, low-frequency components representing illumination, mid-frequency components capturing texture, and high-frequency components often corresponding to noise and fine details.

\subsubsection{Local Discrete Cosine Transform (LDCT) descriptor}\label{subsection:LDCT}
The LDCT divides the thyroid nodule patch into small, non-overlapping square cells and applies the DCT to each cell independently. If $M$ represents the height and width of a cell, where $M \ll N$, then \(  \mathcal{C}_{i,j} \in \mathbb{R}^{M \times M} \) represents the local cell at the $i^{th}$ row and $j^{th}$ column of this patch. Thus, we obtain \( \frac{N}{M} \times \frac{N}{M} \) non-overlapping cells and the thyroid nodule patch is represented as (\cite{singh2007optimization})

\begin{equation}
	\mathtt{I} = \bigcup_{i=1}^{\frac{N}{M}} \bigcup_{j=1}^{\frac{N}{M}} \mathcal{C}_{i,j} .
	\label{eq:cell_partition}
\end{equation}

Given the intensity of a pixel in a cell \( C_{i,j}(x',y') \), where  \( x' \in \{0,1,\ldots,M-1\} \) and \( y' \in \{0,1,\ldots,M-1\} \) denote the spatial coordinates within the cell,  the frequency representation of the cell obtained by applying the DCT is given as follows:

\begin{equation}
	{F}_{i,j}(u', v') = \alpha(u')\alpha(v') \sum_{x'=0}^{M-1} \sum_{y'=0}^{M-1} {C}_{i,j}(x', y')
	\cos\left[\frac{(2x'+1)u'\pi}{2M}\right] \cos\left[\frac{(2y'+1)v'\pi}{2M}\right],
	\label{eq:local_dct}
\end{equation}
\noindent for \( u', v' = 0, 1, \ldots, M-1 \), and normalization term \( \alpha(\cdot) \) defined as in Eq.~\eqref{eq:normalization} with $N$ replaced by $M$.

In this frequency matrix, the element in the first row and the first column represents the average brightness of the cell \( \mathcal{C}_{i,j} \) and is termed as the DC coefficient. The top left corner of this matrix carries the low-frequency coefficients, which, as before, represents the overall lighting condition. The bottom right elements of this matrix carry the high-frequency coefficients and, as mentioned before, represent the distortion and noise. Further, in between elements of the matrix are the middle-frequency coefficients, which represent the textural features. All these elements are termed as the AC coefficients.

\begin{figure*}[!h]
	\centerline{\includegraphics[width=0.45\columnwidth]{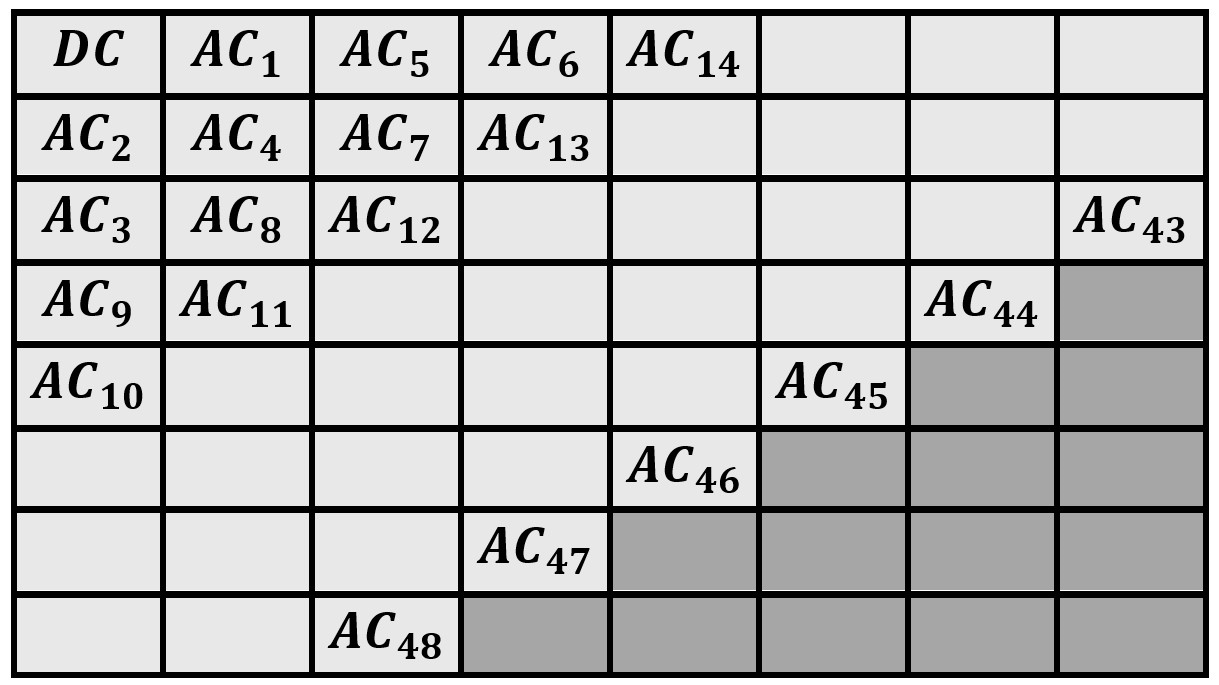}}
	\caption{Zigzag selection of DCT coefficients of cell \( {C}_{i,j} \). }
	\label{fig: cell_coeff}
\end{figure*}

In general, ultrasound images contain noise (\cite{morin2015motion}). This is true for thyroid nodule patches as well. Therefore, we discard  high-frequency coefficients from each cell and only retain low and middle-frequency coefficients along with the DC coefficient, as most of the studies have done (\cite{shastri2018density, dabbaghchian2010feature}). These coefficients are selected in a conventional zigzag manner as shown in Figure \ref{fig: cell_coeff} (\cite{agrawal2022sabmis, acharya2016integrated}). The selected coefficients are indexed as \( \ell = 0, 1, \ldots, L-1 \).

\subsubsection{Improved Local Binary Pattern (ILBP) descriptor}{\label{subsubsection:ILBP}
The ILBP is a simple technique used to describe the texture of an image by looking at small patterns around each pixel. Every pixel is surrounded by a \(3\times 3\) block. If we choose the corners of this block and the midpoints of the lines connecting the corners, then we have a total of 8 neighbors. Now, the intensity of each neighbor is compared with the mean intensity of all the pixels in the block.  If the pixel intensity of the neighbor is greater than or equal to the mean intensity value, it is assigned a value of 1; otherwise, it is assigned 0. These binary values are arranged in a clockwise direction starting from the top-left neighbor, forming an 8-bit binary number. This number is then converted into a decimal value, which becomes the new value for the center pixel as depicted in Figure \ref{fig: ILBP_img}.
\begin{figure*}[!h]
	\centerline{\includegraphics[width=0.50\columnwidth]{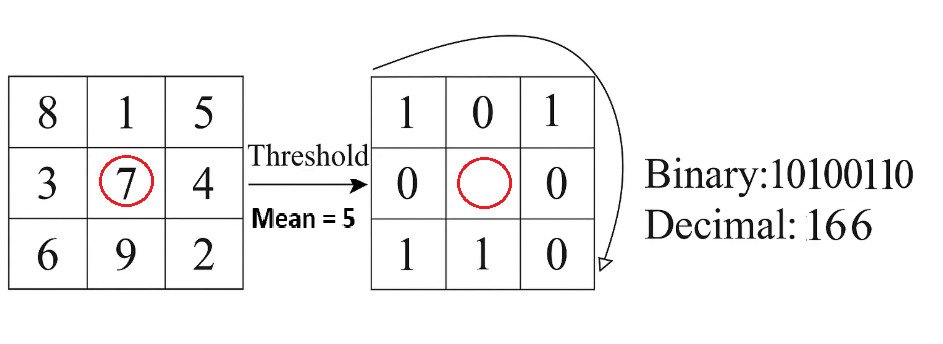}}
	\caption{Binary pattern conversion using ILBP. }
	\label{fig: ILBP_img}
\end{figure*} \\
This can be mathematically represented as follows \cite{huang2011local}:
\begin{equation}
	\text{ILBP}_c= \sum_{p=0}^{7} s(i_p - \bar{i}) \cdot 2^p,
	\label{eq:ilbp}
\end{equation}
where \(i_p\) is the intensity of the \(p^\text{th}\) pixel in the neighborhood, and \(\bar{i}\) is the mean intensity of all pixels in the block, and the function $s(z)$ is defined as follows:

\begin{equation}
	s(z) =
	\begin{cases}
			1, & \text{if } z \geq 0 \\
			0, & \text{otherwise}
		\end{cases}.
	\label{eq:lbp_2}
\end{equation}
Repeating this for every pixel in the thyroid nodule image, the ILBP descriptor creates a new image where each pixel represents a pattern of texture in its local area.

%\textcolor{blue}{We now present the functioning of our proposed novel LDBP descriptor, which demonstrates how ILBP is applied to the Local DCT coefficients. Earlier, the ILBP has been applied to the spatial domain. We are the first to apply it to the frequency domain.}

\subsubsection{Binary Pattern Driven Local Discrete Cosine Transform  (BPD-LDCT)  Descriptor}{\label{subsubsection:BPDLDCT}}

%As discussed above, we have $\frac{N}{M} \times \frac{N}{M}$ cells and in each cell the \( \ell = 0, 1, \ldots, L-1 \) represents the selected coefficients.

In BPD-LDCT, instead of applying ILBP on each pixel intensity we apply it on the frequency coefficients of each cell obtained from LDCT. Recall, we have $\frac{N}{M} \times \frac{N}{M}$ cells in a thyroid nodule patch.
The binary pattern corresponding to the $\ell^{th}$ coefficient of the ($i$, $j$) cell of the thyroid nodule patch is given as follows:
\begin{equation}
B_{i,j}(\ell) = s\left(c_{i,j}^{\ell} - \mu_{i,j}\right) \quad  \forall \ell = 0, 1, \ldots, L-1 ,
\end{equation}
where \( c_{i,j}^{\ell} \) is the coefficient at index $\ell$ of the ($i,j$) cell, \( \mu_{i,j} \)  represents the mean of all the selected coefficients in ($i,j$) cell, and \( s(.) \) is the threshold function, and is calculated as in Eq \ref{eq:lbp_2}.
Next, this binary pattern is converted into decimal form as
\begin{equation}
	D_{i,j}(\ell) = B_{i,j}(\ell) \cdot 2^\ell, \quad  \forall \ell = 0, 1, \ldots, L-1 .
\end{equation}
The final descriptor for the ($i, j$) cell of the thyroid nodule patch as follows:

\begin{equation}
	\text{BPD-LDCT}_{i,j} = \sum_{\ell=0}^{L} D_{i,j}(\ell).
\end{equation}
This process is repeated for all cells in the thyroid nodule patch. Finally, the extracted features are fed to our next component, which is data balancing, to balance the samples.

\subsection{Data Balancing} \label{subsection : Smote}

Both the datasets, TDID and AUITD, are imbalanced, i.e., the benign class (the minority class) has significantly fewer samples than the malignant class (the majority class). At this stage of our model processing, a sample yields an encoded feature vector. This imbalance can lead to biased model performance, where the model becomes more accurate at predicting the majority class as compared to the minority class.

The most common approach to address this issue is by minority oversampling. This increases the number of samples of the minority class by generating new synthetic data samples. Thus, the number of samples in the minority class becomes equal to that of the majority class. For this study, we use one of the most common minority oversampling techniques, namely SMOTE (\cite{liu2024feature, chawla2002smote}). It generates synthetic samples for the minority class. Next, the balanced samples are fed into our classifier below.

\subsection{Classification} \label{subsection : classification}
In our proposed technique, we use a non-linear SVM using an RBF kernel to capture complex, non-linear relationships (\cite{ zouhri2022handling, DONASCIMENTO201865}). As mentioned earlier, the classification is done in two stages. In Stage $\mathrm{I}$, the SVM classifies the thyroid patch as either benign or malignant. In Stage $\mathrm{II}$, malignant images are further classified based on their TI-RADS score into TI-RADS (4) or TI-RADS (5) categories.

\section{Dataset and experimental results}\label{section: result}
This section is divided into three subsections. Section \ref{E_M} describes the experimental settings and evaluation metrics. Section \ref{s1} presents the datasets used for Stage $\mathrm{I}$ classification along with the corresponding numerical results. Section \ref{s2} covers the datasets used for Stage $\mathrm{II}$ classification and their respective results. Next, we describe the hyper-parameters used in our experiments. Multiple values of these hyperparametes were experimented with and the best ones are reported. The value of M, which denotes the height and width of a cell, is taken as $8$.  The value of $L$, which represents the number of selected frequency coefficients in a cell, is taken as $36$. The implementation a conducted in Python, using established libraries including NumPy, OpenCV, Scikit-learn, Pandas, Matplotlib, etc.

\subsection{Experimental settings \& evaluation metrics}\label{E_M}
 We use a \textit{K-fold} cross-validation approach for training and testing, where the value of $K$ is $5$, which provides a good balance between computational efficiency and model performance estimation (\cite{marcot2021optimal, rastogi2024multi}) . In 5-fold cross-validation, the data is randomly split into five equal (or nearly equal) parts, called folds. The model is trained and tested five times, each using a different fold as the test set and the remaining four as the training set. After completing five iterations, the performance metrics from each iteration are averaged to produce a final result. This gives reliable and unbiased estimates of the model performance on unseen data. 

To assess the performance of our system, we use several standard metrics such as specificity, sensitivity, precision, F1-score, and accuracy. Specificity (\textit{Spec}) assesses the model's ability to correctly identify negative instances out of the total number of actual negatives. It is calculated as follows:

\begin{equation}
	\textit{Spec} = \frac{TN}{TN + FP},
\end{equation}
where TN and FP mean True Negative and False Positive, respectively.

Sensitivity (\textit{Sens}) measures the model's ability to correctly identify positive instances out of the total actual positives. It is calculated as follows:

\begin{equation}
	\textit{Sens} = \frac{TP}{TP + FN},
\end{equation}
where TP and FN mean True Positive and False Negative, respectively.

Precision (\textit{Pre}) of the model is the amount of correctly identified positive cases compared to the total number of predicted positives. It is computed as follows:
\begin{equation}
	%\scriptsize
	\begin{aligned}
		\textit{Pre} = \frac{TP}{TP + FP }.\\
	\end{aligned}
\end{equation}

F1-Score (\textit{F1}) is determined by taking the harmonic mean of recall and precision and \mcs{is} mathematically calculated as follows:
\begin{equation}
	%\scriptsize
	\begin{aligned}
		\textit{F1} = 2 \times \frac{Precision \times Sensitivity}{Precision + Sensitivity}.\\
	\end{aligned}
\end{equation}

Accuracy (\textit{Acc}) represents the overall correctness of the classification model and is calculated as the ratio of correctly predicted instances to the total instances.
\begin{equation}
	%\scriptsize
	\begin{aligned}
		\textit{Acc} = \frac{TP + TN}{TP + FN +  TN + FP }.\\
	\end{aligned}
\end{equation}

\subsection{Stage $\mathrm{I}$ classification}\label{s1}
This subsection is organized into two parts. Section \ref{ds1} describes the dataset distribution for Stage $\mathrm{I}$, Section \ref{rs1_no_aug} presents the numerical results on both the datasets.

\subsubsection{Datasets for Stage $\mathrm{I}$ classification}\label{ds1}

To evaluate the robustness of our proposed method, we use two publicly available datasets, namely Thyroid Digital Image Dataset (TDID) (\cite{pedraza2015open}) and Algerian Ultrasound Images Thyroid Dataset (AUITD) (\cite{azouz2023}). Both datasets have been annotated by medical experts, as reported in the original dataset publications, and the images are categorized as either benign or malignant. The TDID dataset contains a total of $349$ images, of which $61$ are benign and $288$ are malignant. Each image in the TDID dataset has a resolution of $560\times360$ pixels, and we resize them without cropping to a size of $256\times256$ pixels. In the AUITD dataset, not all images are properly captured (many have manual noise around the nodule area), and only $781$ images are clean. Out of these, $182$ are benign and $599$ are malignant. The images in the AUITD dataset have different resolutions, and we resize them using the bilinear interpolation to a size of $256\times256$ pixels for consistency. Table~\ref{tab:dataset_1} presents the counts of benign and malignant cases for the TDID and AUITD datasets.

%Figure \ref{fig: sample} depicts the sample images of thyroid nodules for each class from the TDID dataset. The images from the AUTID dataset are similar.

\begin{table*}[!h]
	\centering

	\vspace{2pt} % <-- Optional vertical space

	\caption{Distribution of original images across benign and malignant categories for Stage $\mathrm{I}$ classification.}
	\vspace{5pt}

	\resizebox{.4\textwidth}{!}{
		\begin{tabular}{c|c|c|c}
			\hline
			\multirow{2}{*}{Dataset} & \multirow{2}{*}{Category} & \multicolumn{2}{c}{Number of Images} \\
			\cline{3-4}
			& & Original & Total \\
			\hline
			\multirow{2}{*}{TDID} & Benign    & 61  & \multirow{2}{*}{349} \\
			& Malignant & 288 &                      \\
			\hline
			\multirow{2}{*}{AUITD} & Benign    & 182 & \multirow{2}{*}{781} \\
			& Malignant & 599 &                      \\
			\hline
		\end{tabular}
	}

	\vspace{2pt} % <-- Optional vertical space

	\label{tab:dataset_1}
\end{table*}

\subsubsection{Numerical results for Stage $\mathrm{I}$}\label{rs1_no_aug}
The focus of this subsection is to compare results of different algorithms for Stage $\mathrm{I}$ classification on both the datasets. Table \ref{tab:result_comparison_no_augmentation} shows how DCT, LBP, and several state-of-the-art techniques performed as compared to our proposed BPD-LDCT descriptor. Figure~\ref{fig:stage1_marker_metrics} provides the corresponding marker-based visual comparison. For the remainder of this discussion, the comparison is based on the average performance of all available techniques.

On the TDID dataset, DCT achieves a performance of $97.82\%$, LDCT achieves $98.25\%$, ILBP achieves $89.23\%$, and the best method among the existing state-of-the-art studies achieves $95.00\%$. In contrast, our BPD-LDCT descriptor achieves an excellent performance of  $\textbf{99.59\%}$, clearly outperforming all others. For the AUITD dataset, DCT records $91.13\%$, LDCT records $92.23\%$ and ILBP records $86.35\%$, with no prior works available for direct comparison. The proposed BPD-LDCT again records a strong average performance of $\textbf{96.86\%}$, surpassing all three.

The strong performance of our proposed BPD-LDCT descriptor highlights the effectiveness in capturing rich and meaningful texture features that are often missed by traditional techniques.

\begin{table*}[!h]
	%\scriptsize
	\centering
	\caption{Quantitative comparison of the proposed BPD-LDCT descriptor with standard methods and with existing state-of-the-art techniques for Stage $\mathrm{I}$ classification on both TDID and AUITD datasets.}
	\vspace{0.3cm}
	\resizebox{\columnwidth}{!}{%
		\begin{tabular}{l|l|c|c|c|c|c|c}
			\hline
			%& &  &   &  &  & & \\
			Datasets & Technique  & \textit{Pre (\%)} & \textit{F1 (\%)} & \textit{Spec (\%)} &  \textit{Sen (\%)} & \textit{Acc (\%)} & \textit{Avg (\%)}\\ \hline
			%& &  &   &  &  & & \\ \hline

			%&&&&&&&\\
			\multirow{10}{*}{TDID}& DCT   &97.42 &  97.83 & 97.68  & 98.29  & 97.91 &97.82\\
			%&&&&&&&\\
			&LDCT & 98.17 &  98.22 & 98.31 & 98.29 & 98.26 &98.25\\
			%&&&&&&&\\
			&ILBP &90.14 &88.73&90.71&87.53&89.05&89.23\\  \cline{2-8}
			%& &  &   &  &  & & \\
			&MC-CNN (\cite{song2018multitask})  & \textbf{-}& \textbf{-}&   96.20   &   94.10 &   92.10 & 94.13\\
			%&&&&&&&\\
			& FFT, ResNet18/34/50 (\cite{nguyen2019artificial})&\textbf{-} & \textbf{-}&   63.74   & 94.93  &  90.88  &83.18\\
			%&&&&&&&\\
			&ResNet50 + Inception (\cite{nguyen2020ultrasound}) &\textbf{-} &\textbf{-} & 65.68  &  96.07  &  92.05 &84.60\\
			%&&&&&&&\\

			&SURF + ResNet18 (\cite{hang2021thyroid})   &\textbf{-} &  \textbf{-}  &  \textbf{-} &   \textbf{-} & 95.00& 95.00\\
			%&&&&&&&\\
			&Deep-GAN (\cite{srivastava2024deep})     & \textbf{-} &  95.65    &   94.91      &   96.70     &  96.00& 95.81\\ \cline{2-8}
			%&&&&&&&\\
			& \textbf{BPD-LDCT (Proposed) }     &\textbf{99.36} &  \textbf{99.67 }         & \textbf{99.25} & \textbf{100}  & \textbf{99.65}         &\textbf{99.59} \\
			%&&&&&&& \\
			\hline \hline

			%&&&&&&&\\

			\multirow{4}{*}{AUITD}& DCT  & 90.72 & 91.31  & 90.45   & 91.98 & 91.23      &   91.13\\
			%&&&&&&&\\
			&LDCT & 91.90 &  92.36 & 91.62 & 92.25 & 92.32 &92.23\\
			%&&&&&&&\\
			&ILBP &87.18 &86.04&87.42&84.90&86.22&86.35\\  \cline{2-8}
			%& &  &   &  &  & & \\
			& \textbf{BPD-LDCT (Proposed) }        &  \textbf{96.38} &  \textbf{96.99}          & \textbf{96.32}   & \textbf{97.63}  & \textbf{96.99}    & \textbf{96.86}  \\
			%&&&&&&&\\
			\hline

	\end{tabular}}
	\label{tab:result_comparison_no_augmentation}

\end{table*}

\begin{figure}[!h]
	\centering
	\begin{subfigure}[b]{0.49\textwidth}
		\centering
		\begin{tikzpicture}
			\begin{axis}[
				width=\textwidth,
				height=6cm,
				xtick=data,
				xticklabels={\textit{Pre}, \textit{F1}, \textit{Spec},\textit{ Sen}, \textit{Acc}, \textit{Avg}},
				ylabel={\scriptsize Score (\%)},
				ymin=70, ymax=101,
				ytick={75,80,85,90,95,100},
				xticklabel style={font=\scriptsize},
				yticklabel style={font=\scriptsize},
				legend style={font=\tiny, at={(0.5,0.15)}, anchor=north, legend columns=4, /tikz/every even column/.append style={column sep=0.4cm}},
				grid=major,
				scatter/classes={
					DCT={mark=*,blue},
					LDCT={mark=*,olive},
					ILBP={mark=*,red},
					BPD-LDCT={mark=*,black!60!green}
				},
				scatter
				]
				% TDID dataset
				\addplot[scatter, only marks, scatter src=explicit symbolic]
				coordinates {
					(0,97.42) [DCT]
					(1,97.83) [DCT]
					(2,97.68) [DCT]
					(3,98.29) [DCT]
					(4,97.91) [DCT]
					(5,97.82) [DCT]
				};
				\addplot[scatter, only marks, scatter src=explicit symbolic]
				coordinates {
					(0,90.14) [ILBP]
					(1,88.73) [ILBP]
					(2,90.71) [ILBP]
					(3,87.53) [ILBP]
					(4,89.05) [ILBP]
					(5,89.23) [ILBP]
				};
				\addplot[scatter, only marks, scatter src=explicit symbolic]
				coordinates {
					(0,98.17) [LDCT]
					(1,98.22) [LDCT]
					(2,98.31) [LDCT]
					(3,98.29) [LDCT]
					(4,98.26) [LDCT]
					(5,98.25) [LDCT]
				};

				\addplot[scatter, only marks, scatter src=explicit symbolic]
				coordinates {
					(0,99.36) [BPD-LDCT]
					(1,99.67) [BPD-LDCT]
					(2,99.25) [BPD-LDCT]
					(3,100.0) [BPD-LDCT]
					(4,99.65) [BPD-LDCT]
					(5,99.59) [BPD-LDCT]
				};
				\legend{DCT, LDCT, ILBP,  BPD-LDCT (Proposed)}
			\end{axis}
		\end{tikzpicture}
		\caption{Comparison of our method with standard methods on the TDID dataset}
	\end{subfigure}
	\hfill
		\begin{subfigure}[b]{0.49\textwidth}
			\centering
			\begin{tikzpicture}
				\begin{axis}[
					width=\textwidth,
					height=6cm,
					xtick=data,
					xticklabels={\textit{Pre}, \textit{F1}, \textit{Spec},\textit{ Sen}, \textit{Acc}, \textit{Avg}},
					ylabel={\scriptsize Score (\%)},
					ymin=50, ymax=101,
					ytick={60, 65,70,75,80,85,90,95,100},
					xticklabel style={font=\scriptsize},
					yticklabel style={font=\scriptsize},
					legend style={font=\tiny, at={(0.5,0.2)}, anchor=north, legend columns=3, /tikz/every even column/.append style={column sep=0.25cm}},
					grid=major,
					scatter/classes={
						FFTResNet={mark=*,brown},
						ResIncep={mark=*,gray},
						MCCNN={mark=*,cyan!60!black},
						SURFRes={mark=*,magenta!60!black},
						DeepGAN={mark=*,orange},
						BPD-LDCT={mark=*,black!60!green}
					},
					scatter
					]
					\addplot[scatter, only marks, scatter src=explicit symbolic]
					coordinates {
						(0,99.36) [BPD-LDCT]
						(1,99.67) [BPD-LDCT]
						(2,99.25) [BPD-LDCT]
						(3,100.0) [BPD-LDCT]
						(4,99.65) [BPD-LDCT]
						(5,99.59) [BPD-LDCT]
					};
					\addplot[scatter, only marks, scatter src=explicit symbolic]
					coordinates {
						(1,95.65) [DeepGAN]
						(2,94.91) [DeepGAN]
						(3,96.70) [DeepGAN]
						(4,96.00) [DeepGAN]
						(5,95.81) [DeepGAN]
					};
					% BPD-LDCT
					% FFT+ResNet
					\addplot[scatter, only marks, scatter src=explicit symbolic]
					coordinates {
						(2,63.74) [FFTResNet]
						(3,94.93) [FFTResNet]
						(4,90.88) [FFTResNet]
						(5,83.18) [FFTResNet]
					};
					% ResNet+Incep.
					\addplot[scatter, only marks, scatter src=explicit symbolic]
					coordinates {
						(2,65.68) [ResIncep]
						(3,96.07) [ResIncep]
						(4,92.05) [ResIncep]
						(5,84.60) [ResIncep]
					};
					% MC-CNN
					\addplot[scatter, only marks, scatter src=explicit symbolic]
					coordinates {
						(2,96.20) [MCCNN]
						(3,94.10) [MCCNN]
						(4,92.10) [MCCNN]
						(5,94.13) [MCCNN]
					};
					% SURF+ResNet
					\addplot[scatter, only marks, scatter src=explicit symbolic]
					coordinates {
						(4,95.00) [SURFRes]
						(5,95.00) [SURFRes]
					};
					% Deep-GAN

					\legend{FFT+ResNet, ResNet+Incep., MC-CNN, SURF+ResNet, Deep-GAN, BPD-LDCT (Proposed)}
				\end{axis}
			\end{tikzpicture}
			\caption{Comparison of our method with state-of-the-art methods on the TDID dataset}
		\end{subfigure}

	\hfill

	\begin{subfigure}[b]{0.49\textwidth}
		\centering
		\begin{tikzpicture}
			\begin{axis}[
				width=\textwidth,
				height=6cm,
				xtick=data,
				xticklabels={\textit{Pre}, \textit{F1}, \textit{Spec},\textit{ Sen}, \textit{Acc}, \textit{Avg}},
				ylabel={\scriptsize Score (\%)},
				ymin=70, ymax=101,
				ytick={75,80,85,90,95,100},
				xticklabel style={font=\scriptsize},
				yticklabel style={font=\scriptsize},
				legend style={font=\tiny, at={(0.5,0.15)}, anchor=north, legend columns=4, /tikz/every even column/.append style={column sep=0.4cm}},
				grid=major,
				scatter/classes={
					DCT={mark=*,blue},
					LDCT={mark=*,olive},
					ILBP={mark=*,red},
					BPD-LDCT={mark=*,black!60!green}
				},
				scatter
				]
					% AUITD dataset
				\addplot[scatter, only marks, scatter src=explicit symbolic]
				coordinates {
					(0,90.72) [DCT]
					(1,91.31) [DCT]
					(2,90.45) [DCT]
					(3,91.98) [DCT]
					(4,91.23) [DCT]
					(5,91.13) [DCT]
				};

				\addplot[scatter, only marks, scatter src=explicit symbolic]
				coordinates {
					(0,91.90) [LDCT]
					(1,92.36) [LDCT]
					(2,91.62) [LDCT]
					(3,92.25) [LDCT]
					(4,92.32) [LDCT]
					(5,92.32) [LDCT]
				};

				\addplot[scatter, only marks, scatter src=explicit symbolic]
				coordinates {
					(0,87.18) [ILBP]
					(1,86.04) [ILBP]
					(2,87.42) [ILBP]
					(3,84.90) [ILBP]
					(4,86.22) [ILBP]
					(5,86.35) [ILBP]
				};

				\addplot[scatter, only marks, scatter src=explicit symbolic]
				coordinates {
					(0,96.38) [BPD-LDCT]
					(1,96.99) [BPD-LDCT]
					(2,96.32) [BPD-LDCT]
					(3,97.63) [BPD-LDCT]
					(4,96.99) [BPD-LDCT]
					(5,96.86) [BPD-LDCT]
				};
				\legend{DCT, LDCT, ILBP,  BPD-LDCT (Proposed)}
			\end{axis}
		\end{tikzpicture}
		\caption{Comparison on the AUITD dataset}
	\end{subfigure}

	\caption{Evaluation metric comparison for Stage $\mathrm{I}$ classification using marker-only plots. Each point represents the score of a specific evaluation metric achieved by different techniques.}
	\label{fig:stage1_marker_metrics}
\end{figure}
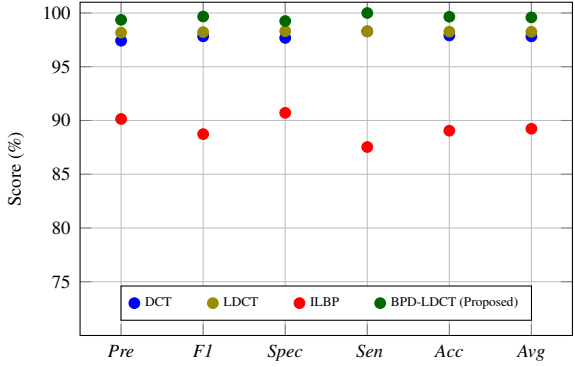
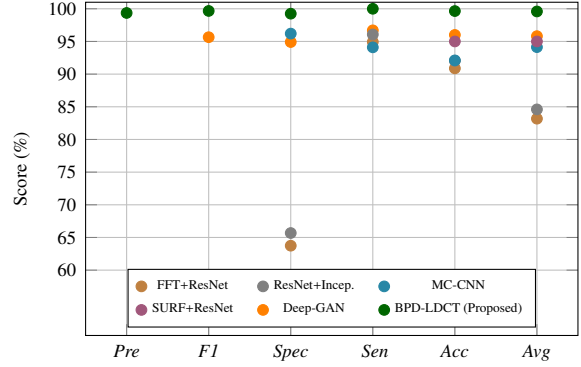
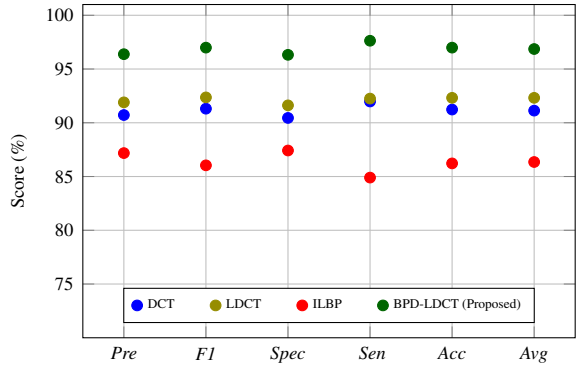

\subsection{Stage $\mathrm{II}$ classification}\label{s2}
This subsection is divided into two parts. Section \ref{ds2} outlines the dataset distribution for stage $\mathrm{II}$, Section \ref{rs2_no_aug} presents the numerical results on both the datasets.
\subsubsection{Dataset for Stage $\mathrm{II}$ classification}\label{ds2}

In both the datasets, benign and malignant images are further labelled based upon their TI-RADS (Thyroid Imaging Reporting and Data System) score, which includes scores ``2", ``3", ``4a", ``4b", ``4c", and ``5". Here, TI-RADS score ``2" represents benign nodules; TI-RADS score ``3" indicates nodules that are probably benign with no ultrasound features; TI-RADS score ``4a" corresponds to a low apprehension of malignancy with one suspicious ultrasound feature; TI-RADS score ``4b" corresponds to an intermediate apprehension of malignancy with two suspicious features; and TI-RADS category ``4c" corresponds to a moderate apprehension of malignancy with three or four suspicious features; TI-RADS category ``5" signifies a high apprehension of malignancy with five suspicious ultrasound features.

In Stage $\mathrm{II}$, benign cases are not classified any further, as they generally do not require immediate medical intervention or intensive follow-up. Malignant cases, on the other hand, are further classified into TI-RADS (4) and TI-RADS (5) categories. This categorization is clinically significant, as it helps in assessing the level of risk and guiding appropriate treatment strategies. For example, in one study the risk of TI-RADS (5) cancer was twice that of TI-RADS (4) cancer (\cite{mohanty2019role}). Table \ref{tab:dataset_2} shows the distribution of malignant cases across TI-RADS (4) and TI-RADS (5) categories.

\begin{table*}[!h]
	\centering
	\caption{Distribution of malignant images across TI-RADS (4) and TI-RADS (5) categories for Stage $\mathrm{II}$ classification.}
	\vspace{5pt}
	\resizebox{.4\textwidth}{!}{
		\begin{tabular}{c|c|c|c}
			\hline
			\multirow{2}{*}{Dataset} & \multirow{2}{*}{Category} & \multicolumn{2}{c}{Number of Images} \\
			\cline{3-4}
			& & Original & Total \\
			\hline
			\multirow{2}{*}{TDID}  & TI-RADS (4) & 243 & \multirow{2}{*}{288} \\
			& TI-RADS (5) & 45  &                      \\
			\hline
			\multirow{2}{*}{AUITD} & TI-RADS (4) & 505 & \multirow{2}{*}{599} \\
			& TI-RADS (5) & 94  &                      \\
			\hline
		\end{tabular}
	}
	\label{tab:dataset_2}
\end{table*}

%\begin{figure*}[!h]
%	\centerline{\includegraphics[width=0.55\columnwidth]{images/Fig7.png}}
%	\caption{\scriptsize Sample thyroid nodule images in the TDID dataset \cite{pedraza2015open}}
%	\label{fig: sample}
%\end{figure*}

\subsubsection{Numerical results for Stage $\mathrm{II}$}\label{rs2_no_aug}
This subsection aims to compare the performance of different algorithms for Stage $\mathrm{II}$ classification across both datasets. To the best of our knowledge, this is the first study to address Stage $\mathrm{II}$ classification, where malignant nodules are categorized based on their TI-RADS scores. As no prior work exists for direct comparison, we evaluate our proposed BPD-LDCT descriptor against DCT, LDCT, and ILBP, across both the datasets. Table \ref{tab:stage_2} presents the numerical results, and Figure~\ref{fig:ldbplineplots_stage2_noaug} depicts the corresponding marker plots.

On the TDID dataset, DCT attains a performances of $97.30\%$, LDCT attains $98.26\%$, and ILBP attains $91.51\%$. In comparison, our proposed BPD-LDCT descriptor attains an average performance of $\textbf{99.52\%}$. On the AUITD dataset, DCT deliers a performances of $97.92\%$, LDCT delivers $98.38\%$, and ILBP delivers $94.77\%$. Relative to these results, our proposed BPD-LDCT descriptor records an average performance of $\textbf{99.29\%}$.

Overall, these results clearly demonstrate the superiority of the proposed BPD-LDCT descriptor in Stage $\mathrm{II}$ classification, consistently outperforming traditional texture descriptors across both the datasets.

\begin{table*}[!h]
	%\scriptsize
	\centering
	\caption{Quantitative comparison of the proposed BPD-LDCT descriptor with standard methods for Stage $\mathrm{II}$ classification on both TDID and AUITD datasets.}

	\vspace{0.3cm}
	\resizebox{0.85\columnwidth}{!}{%
		\begin{tabular}{l|l|c|c|c|c|c|c}
			\hline
			%& &  &   &  &  & & \\
			Datasets & Technique  & \textit{Pre (\%)} & \textit{F1 (\%)} & \textit{Spec (\%)} &  \textit{Sen (\%)} & \textit{Acc (\%)} & \textit{Avg (\%)}\\ \hline
			%& &  &   &  &  & & \\ \hline
			%& &  &   &  &  & & \\ \hline
			%& &  &   &  &  & & \\
			\multirow{5}{*}{TDID} & DCT  &97.92 & 97.12  & 97.95  & 96.39 & 97.11   &  97.30    \\
			%& &  &   &  &  & & \\
			&LDCT & 98.75 &  98.14 & 98.75 & 97.55& 98.14&98.26\\
			%& &  &   &  &  & & \\
			&ILBP & 90.39 &91.79   & 90.23& 93.40 &91.76 &91.51 \\
			%& &  &   &  &  & & \\
			& \textbf{BPD-LDCT (Proposed) }       &  \textbf{100} &  \textbf{99.40}          & \textbf{100}   & \textbf{98.81}  & \textbf{99.38}     & \textbf{99.52} \\
			%& &  &   &  &  & & \\
			\hline
			%& &  &   &  &  &  &\\

			\multirow{5}{*}{AUITD}& DCT   & 98.78& 97.67 & 98.81  &96.60 & 97.72   &  97.92  \\
			%& &  &   &  &  & & \\
			&LDCT & 99.00 &  98.22 & 99.01 & 97.48 & 98.21 &98.38\\
			%& &  &   &  &  & & \\
			&ILBP &95.36  &  94.54 &95.49 &  93.81&94.65 & 94.77\\
			%& &  &   &  &  & & \\
			& \textbf{BPD-LDCT (Proposed) }       &  \textbf{99.61} &  \textbf{99.22}          & \textbf{99.58}   & \textbf{98.84}  & \textbf{99.20}    & \textbf{99.29}  \\
			%& &  &   &  &  &&  \\
			\hline

	\end{tabular}}

	\label{tab:stage_2}

\end{table*}

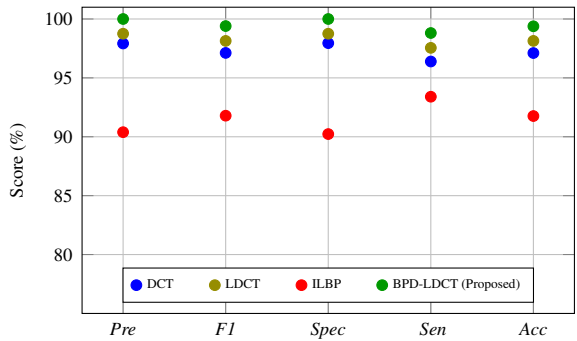
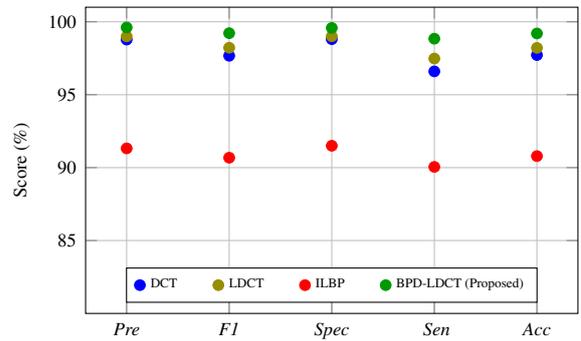
\begin{figure}[!h]
	\centering
	\begin{subfigure}[b]{0.49\textwidth}
		\centering
		\begin{tikzpicture}
			\begin{axis}[
				width=\textwidth,
				height=0.7\textwidth,
				xtick=data,
				xticklabels={\textit{Pre}, \textit{F1}, \textit{Spec},\textit{ Sen}, \textit{Acc}, \textit{Avg}},
				ylabel={\scriptsize Score (\%)},
				ymin=75, ymax=101,
				ytick={80, 85, 90, 95, 100},
				grid=major,
				enlarge x limits=0.1,
				tick label style={font=\scriptsize},
				label style={font=\scriptsize},
				legend style={
					at={(0.5,0.15)},
					anchor=north,
					legend columns=4,
					font=\tiny,
					/tikz/every even column/.append style={column sep=0.4cm}
				},
				mark size=2pt,
				]
				\addplot[color=blue, mark=*, only marks] coordinates {
					(0,97.92) (1,97.12) (2,97.95) (3,96.39) (4,97.11)
				};
				\addlegendentry{DCT}

				\addplot[color=olive, mark = *, only marks] coordinates {
					(0,98.75) (1,98.14) (2,98.75) (3,97.55) (4,98.14)
				};
				\addlegendentry{LDCT}

				\addplot[color=red, mark=*, only marks] coordinates {
					(0,90.39) (1,91.79) (2,90.23) (3,93.40) (4,91.76)
				};
				\addlegendentry{ILBP}

				\addplot[color=green!60!black, mark=*, only marks] coordinates {
					(0,100) (1,99.40) (2,100) (3,98.81) (4,99.38)
				};
				\addlegendentry{BPD-LDCT (Proposed)}
			\end{axis}
		\end{tikzpicture}
		\caption{Comparison on the TDID dataset}
	\end{subfigure}
	\hfill
	\begin{subfigure}[b]{0.49\textwidth}
		\centering
		\begin{tikzpicture}
			\begin{axis}[
				width=\textwidth,
				height=0.7\textwidth,
				xtick=data,
				xticklabels={\textit{Pre}, \textit{F1}, \textit{Spec},\textit{ Sen}, \textit{Acc}, \textit{Avg}},
				ylabel={\scriptsize Score (\%)},
				ymin=80, ymax=101,
				ytick={ 85, 90, 95, 100},
				grid=major,
				enlarge x limits=0.1,
				tick label style={font=\scriptsize},
				label style={font=\scriptsize},
				legend style={
					at={(0.5,0.15)},
					anchor=north,
					legend columns=4,
					font=\tiny,
					/tikz/every even column/.append style={column sep=0.4cm}
				},
				mark size=2pt,
				]
				\addplot[color=blue, mark=*, only marks] coordinates {
					(0,98.78) (1,97.67) (2,98.81) (3,96.60) (4,97.72)
				};
				\addlegendentry{DCT}
				\addplot[color=olive, mark=*, only marks] coordinates {
					(0,99.00) (1,98.22) (2,99.01) (3,97.48) (4,98.21)
				};
				\addlegendentry{LDCT}

				\addplot[color=red, mark=*, only marks] coordinates {
					(0,91.32) (1,90.68) (2,91.50) (3,90.05) (4,90.79)
				};
				\addlegendentry{ILBP}

				\addplot[color=green!60!black, mark=*, only marks] coordinates {
					(0,99.61) (1,99.22) (2,99.58) (3,98.84) (4,99.20)
				};
				\addlegendentry{BPD-LDCT (Proposed)}
			\end{axis}
		\end{tikzpicture}
		\caption{Comparison on the AUITD dataset}
	\end{subfigure}

	\caption{Evaluation metric comparison for Stage $\mathrm{II}$ classification using marker-only plots. Each point represents the score of a specific evaluation metric achieved by different techniques.}
	\label{fig:ldbplineplots_stage2_noaug}
\end{figure}

\section{Conclusion and future work}\label{section:conclusion}
In this study, we present a new CAD system for accurate thyroid cancer classification with a primary focus on feature extraction. Prior studies have highlighted the importance of texture in distinguishing thyroid ultrasound images across different classes. \textit{First}, we conjuncture that this is critical, and based upon our previous work on breast cancer classification, we propose the use of the Discrete Cosine Transform (DCT) descriptor. Thyroid ultrasound images remain challenging since the gland is surrounded by complex anatomical structures that create variations in tissue density. Hence, \textit{second} we conjuncture the need for localization and propose the use of the Local Discrete Cosine Transform (LDCT) descriptor. Furthermore, the intricate thyroid anatomy scatters ultrasound waves, leading to noisy, blurred, and unclear textures. Hence, \textit{third} we conjuncture that a single descriptor cannot fully capture the complete textural information, and we combine another widely-used texture-exploiting descriptor, \mcs{the} Improved Local Binary Pattern (ILBP) with LDCT. ILBP is also recognized for its robustness to noise. We refer to this novel descriptor as the Binary Pattern Driven Local Discrete Cosine Transform (BPD-LDCT). Classification is then performed using a non-linear Support Vector Machine (SVM), yielding robust and accurate diagnostic performance.

We evaluate the effectiveness of the proposed BPD-LDCT descriptor on \textit{the only two publicly available} thyroid datasets, TDID and AUITD, using a two-stage hierarchical classification framework. In Stage $\mathrm{I}$, which distinguishes benign from malignant thyroid nodules, our model achieves a remarkable average performance of nearly $ \mathbf{100\%}$ on TDID, and $\mathbf{97\%}$ on AUITD, surpassing all the existing techniques. In Stage $\mathrm{II}$, which focuses on sub-classifying malignant cases into TI-RADS (4) and TI-RADS (5) categories, our descriptor again attains an exceptional performance of around $ \mathbf{100\%}$ on TDID and $\mathbf{99\%}$ on AUITD.

The future work for this model will focus on several exciting directions. \textit{First}, we plan to enhance the explainability and interpretability of the predictions of the model. This could lead to an automated system that identifies cancerous regions in images more accurately (\cite{saini2024improved}). \textit{Second}, we aim to apply Krylov subspace methods for feature enhancement and regularization, which may lead to improved classification (\cite{CHOUDHARY201856, ahuja2011recycling}). \textit{Third}, we will investigate the application of neural networks in this context, which sometimes works better than SVM that is currently being used (\cite{RAJ2025104910, 9116373}). \textit{Finally}, we will apply this methodology in practice in collaboration with medical doctors.

\section*{CRediT authorship contribution statements}
\textbf{Saurabh Saini:} Conceptualization, Data curation, Investigation, Software, Writing – original draft. \textbf{Kapil Ahuja:} Methodology, Project administration, Resources, Supervision, Writing – review \& editing. \textbf{Marc C. Steinbach:} Formal analysis, Validation, Writing – review \& editing. \textbf{Thomas Wick:} Formal analysis, Validation, Writing – review \& editing.

\section*{Funding}
The authors declare that no funds, grants, or other financial support were received for this study. 

\section*{Competing Interest}
The authors declare that they have no known competing financial interests or personal relationships that could have appeared to influence the work reported in this paper.

\section*{Ethics statements}
All ethical practices have been followed in data usage and publishing.

\bibliographystyle{elsarticle-harv}
\bibliography{bibliography.bib}

\begin{thebibliography}{44}
\expandafter\ifx\csname natexlab\endcsname\relax\def\natexlab#1{#1}\fi
\providecommand{\url}[1]{\texttt{#1}}
\providecommand{\href}[2]{#2}
\providecommand{\path}[1]{#1}
\providecommand{\DOIprefix}{doi:}
\providecommand{\ArXivprefix}{arXiv:}
\providecommand{\URLprefix}{URL: }
\providecommand{\Pubmedprefix}{pmid:}
\providecommand{\doi}[1]{\href{http://dx.doi.org/#1}{\path{#1}}}
\providecommand{\Pubmed}[1]{\href{pmid:#1}{\path{#1}}}
\providecommand{\bibinfo}[2]{#2}
\ifx\xfnm\relax \def\xfnm[#1]{\unskip,\space#1}\fi
%Type = Article
\bibitem[{Acharya et~al.(2016)Acharya, Fujita, Sudarshan, Mookiah, Koh, Tan,
  Hagiwara, Chua, Junnarkar, Vijayananthan et~al.}]{acharya2016integrated}
\bibinfo{author}{Acharya, U.R.}, \bibinfo{author}{Fujita, H.},
  \bibinfo{author}{Sudarshan, V.K.}, \bibinfo{author}{Mookiah, M.R.K.},
  \bibinfo{author}{Koh, J.E.}, \bibinfo{author}{Tan, J.H.},
  \bibinfo{author}{Hagiwara, Y.}, \bibinfo{author}{Chua, C.K.},
  \bibinfo{author}{Junnarkar, S.P.}, \bibinfo{author}{Vijayananthan, A.},
  et~al., \bibinfo{year}{2016}.
\newblock \bibinfo{title}{An integrated index for identification of fatty liver
  disease using {R}adon transform and discrete cosine transform features in
  ultrasound images}.
\newblock \bibinfo{journal}{Information Fusion} \bibinfo{volume}{31},
  \bibinfo{pages}{43--53}.
%Type = Article
\bibitem[{Agrawal et~al.(2022)Agrawal, Ahuja, Steinbach and
  Wick}]{agrawal2022sabmis}
\bibinfo{author}{Agrawal, R.}, \bibinfo{author}{Ahuja, K.},
  \bibinfo{author}{Steinbach, M.C.}, \bibinfo{author}{Wick, T.},
  \bibinfo{year}{2022}.
\newblock \bibinfo{title}{{SABMIS}: Sparse approximation based blind
  multi-image steganography scheme}.
\newblock \bibinfo{journal}{PeerJ Computer Science} \bibinfo{volume}{8},
  \bibinfo{pages}{e1080}.
%Type = Phdthesis
\bibitem[{Ahuja(2011)}]{ahuja2011recycling}
\bibinfo{author}{Ahuja, K.}, \bibinfo{year}{2011}.
\newblock \bibinfo{title}{Recycling Krylov Subspaces and Preconditioners}.
\newblock \bibinfo{type}{Ph.d. thesis}. Virginia Tech.
%Type = Article
\bibitem[{Ali et~al.(2018)Ali, Yasmin, Sharif and Rehmani}]{ALI201839}
\bibinfo{author}{Ali, H.}, \bibinfo{author}{Yasmin, M.},
  \bibinfo{author}{Sharif, M.}, \bibinfo{author}{Rehmani, M.H.},
  \bibinfo{year}{2018}.
\newblock \bibinfo{title}{Computer assisted gastric abnormalities detection
  using hybrid texture descriptors for chromoendoscopy images}.
\newblock \bibinfo{journal}{Computer Methods and Programs in Biomedicine}
  \bibinfo{volume}{157}, \bibinfo{pages}{39--47}.
%Type = Article
\bibitem[{Anand and Gayathri(2015)}]{anand2015mammogram}
\bibinfo{author}{Anand, S.}, \bibinfo{author}{Gayathri, S.},
  \bibinfo{year}{2015}.
\newblock \bibinfo{title}{Mammogram image enhancement by two-stage adaptive
  histogram equalization}.
\newblock \bibinfo{journal}{Optik} \bibinfo{volume}{126},
  \bibinfo{pages}{3150--3152}.
%Type = Misc
\bibitem[{Azouz(2023)}]{azouz2023}
\bibinfo{author}{Azouz, M.}, \bibinfo{year}{2023}.
\newblock \bibinfo{title}{{AUITD}: Algerian ultrasound images thyroid dataset}.
\newblock \URLprefix
  \url{https://www.kaggle.com/datasets/azouzmaroua/algeria-ultrasound-images-thyroid-dataset-auitd}.
  \bibinfo{note}{dataset accessed: 2024-05-01}.
%Type = Article
\bibitem[{Chawla et~al.(2002)Chawla, Bowyer, Hall and
  Kegelmeyer}]{chawla2002smote}
\bibinfo{author}{Chawla, N.V.}, \bibinfo{author}{Bowyer, K.W.},
  \bibinfo{author}{Hall, L.O.}, \bibinfo{author}{Kegelmeyer, W.P.},
  \bibinfo{year}{2002}.
\newblock \bibinfo{title}{{SMOTE}: Synthetic minority over-sampling technique}.
\newblock \bibinfo{journal}{Journal of Artificial Intelligence Research}
  \bibinfo{volume}{16}, \bibinfo{pages}{321--357}.
%Type = Article
\bibitem[{Choudhary and Ahuja(2018)}]{CHOUDHARY201856}
\bibinfo{author}{Choudhary, R.}, \bibinfo{author}{Ahuja, K.},
  \bibinfo{year}{2018}.
\newblock \bibinfo{title}{Stability analysis of bilinear iterative rational
  krylov algorithm}.
\newblock \bibinfo{journal}{Linear Algebra and its Applications}
  \bibinfo{volume}{538}, \bibinfo{pages}{56--88}.
%Type = Article
\bibitem[{Colakoglu et~al.(2019)Colakoglu, Alis and
  Yergin}]{colakoglu2019diagnostic}
\bibinfo{author}{Colakoglu, B.}, \bibinfo{author}{Alis, D.},
  \bibinfo{author}{Yergin, M.}, \bibinfo{year}{2019}.
\newblock \bibinfo{title}{Diagnostic value of machine learning-based
  quantitative texture analysis in differentiating benign and malignant thyroid
  nodules}.
\newblock \bibinfo{journal}{Journal of Oncology} \bibinfo{volume}{2019},
  \bibinfo{pages}{6328329}.
%Type = Article
\bibitem[{Cooper et~al.(2009)Cooper, Doherty, Haugen, Kloos, Lee, Mandel,
  Mazzaferri, McIver, Pacini, Schlumberger et~al.}]{cooper2009revised}
\bibinfo{author}{Cooper, D.S.}, \bibinfo{author}{Doherty, G.M.},
  \bibinfo{author}{Haugen, B.R.}, \bibinfo{author}{Kloos, R.T.},
  \bibinfo{author}{Lee, S.L.}, \bibinfo{author}{Mandel, S.J.},
  \bibinfo{author}{Mazzaferri, E.L.}, \bibinfo{author}{McIver, B.},
  \bibinfo{author}{Pacini, F.}, \bibinfo{author}{Schlumberger, M.}, et~al.,
  \bibinfo{year}{2009}.
\newblock \bibinfo{title}{Revised american thyroid association management
  guidelines for patients with thyroid nodules and differentiated thyroid
  cancer: the {American Thyroid Association (ATA)} guidelines taskforce on
  thyroid nodules and differentiated thyroid cancer}.
\newblock \bibinfo{journal}{Thyroid} \bibinfo{volume}{19},
  \bibinfo{pages}{1167--1214}.
%Type = Article
\bibitem[{Dabbaghchian et~al.(2010)Dabbaghchian, Ghaemmaghami and
  Aghagolzadeh}]{dabbaghchian2010feature}
\bibinfo{author}{Dabbaghchian, S.}, \bibinfo{author}{Ghaemmaghami, M.P.},
  \bibinfo{author}{Aghagolzadeh, A.}, \bibinfo{year}{2010}.
\newblock \bibinfo{title}{Feature extraction using discrete cosine transform
  and discrimination power analysis with a face recognition technology}.
\newblock \bibinfo{journal}{Pattern Recognition} \bibinfo{volume}{43},
  \bibinfo{pages}{1431--1440}.
%Type = Inproceedings
\bibitem[{Dandan et~al.(2018)Dandan, Yakui, Linyao, Xianli and
  Yi}]{dandan2018texture}
\bibinfo{author}{Dandan, L.}, \bibinfo{author}{Yakui, Z.},
  \bibinfo{author}{Linyao, D.}, \bibinfo{author}{Xianli, Z.},
  \bibinfo{author}{Yi, S.}, \bibinfo{year}{2018}.
\newblock \bibinfo{title}{Texture analysis and classification of diffuse
  thyroid diseases based on ultrasound images}, in: \bibinfo{booktitle}{2018
  IEEE International instrumentation and measurement technology conference
  (I2MTC)}, \bibinfo{organization}{IEEE}. pp. \bibinfo{pages}{1--6}.
%Type = Article
\bibitem[{{do Nascimento} et~al.(2018){do Nascimento}, Martins, {Azevedo Tosta}
  and Neves}]{DONASCIMENTO201865}
\bibinfo{author}{{do Nascimento}, M.Z.}, \bibinfo{author}{Martins, A.S.},
  \bibinfo{author}{{Azevedo Tosta}, T.A.}, \bibinfo{author}{Neves, L.A.},
  \bibinfo{year}{2018}.
\newblock \bibinfo{title}{Lymphoma images analysis using morphological and
  non-morphological descriptors for classification}.
\newblock \bibinfo{journal}{Computer Methods and Programs in Biomedicine}
  \bibinfo{volume}{163}, \bibinfo{pages}{65--77}.
%Type = Article
\bibitem[{Ángel E.~Esteban et~al.(2019)Ángel E.~Esteban, López-Pérez,
  Colomer, Sales, Molina and Naranjo}]{ESTEBAN2019303}
\bibinfo{author}{Ángel E.~Esteban}, \bibinfo{author}{López-Pérez, M.},
  \bibinfo{author}{Colomer, A.}, \bibinfo{author}{Sales, M.A.},
  \bibinfo{author}{Molina, R.}, \bibinfo{author}{Naranjo, V.},
  \bibinfo{year}{2019}.
\newblock \bibinfo{title}{A new optical density granulometry-based descriptor
  for the classification of prostate histological images using shallow and deep
  gaussian processes}.
\newblock \bibinfo{journal}{Computer Methods and Programs in Biomedicine}
  \bibinfo{volume}{178}, \bibinfo{pages}{303--317}.
%Type = Article
\bibitem[{Gervasio et~al.(2010)Gervasio, Mujahed, Biasio and
  Alessi}]{gervasio2010ultrasound}
\bibinfo{author}{Gervasio, A.}, \bibinfo{author}{Mujahed, I.},
  \bibinfo{author}{Biasio, A.}, \bibinfo{author}{Alessi, S.},
  \bibinfo{year}{2010}.
\newblock \bibinfo{title}{Ultrasound anatomy of the neck: The infrahyoid
  region}.
\newblock \bibinfo{journal}{Journal of Ultrasound} \bibinfo{volume}{13},
  \bibinfo{pages}{85--89}.
%Type = Article
\bibitem[{Hang(2021)}]{hang2021thyroid}
\bibinfo{author}{Hang, Y.}, \bibinfo{year}{2021}.
\newblock \bibinfo{title}{Thyroid nodule classification in ultrasound images by
  fusion of conventional features and {Res-GAN} deep features}.
\newblock \bibinfo{journal}{Journal of Healthcare Engineering}
  \bibinfo{volume}{2021}, \bibinfo{pages}{9917538}.
%Type = Article
\bibitem[{Huang et~al.(2011)Huang, Shan, Ardabilian, Wang and
  Chen}]{huang2011local}
\bibinfo{author}{Huang, D.}, \bibinfo{author}{Shan, C.},
  \bibinfo{author}{Ardabilian, M.}, \bibinfo{author}{Wang, Y.},
  \bibinfo{author}{Chen, L.}, \bibinfo{year}{2011}.
\newblock \bibinfo{title}{Local binary patterns and its application to facial
  image analysis: a survey}.
\newblock \bibinfo{journal}{IEEE Transactions on Systems, Man, and Cybernetics,
  Part C (Applications and Reviews)} \bibinfo{volume}{41},
  \bibinfo{pages}{765--781}.
%Type = Article
\bibitem[{Jain et~al.(2005)Jain, Nandakumar and Ross}]{jain2005score}
\bibinfo{author}{Jain, A.}, \bibinfo{author}{Nandakumar, K.},
  \bibinfo{author}{Ross, A.}, \bibinfo{year}{2005}.
\newblock \bibinfo{title}{Score normalization in multimodal biometric systems}.
\newblock \bibinfo{journal}{Pattern Recognition} \bibinfo{volume}{38},
  \bibinfo{pages}{2270--2285}.
%Type = Article
\bibitem[{Karanwal(2024)}]{karanwal2024robust}
\bibinfo{author}{Karanwal, S.}, \bibinfo{year}{2024}.
\newblock \bibinfo{title}{Robust face descriptor in unconstrained
  environments}.
\newblock \bibinfo{journal}{Expert Systems with Applications}
  \bibinfo{volume}{247}, \bibinfo{pages}{123302}.
%Type = Article
\bibitem[{Kwak et~al.(2011)Kwak, Han, Yoon, Moon, Son, Park, Jung, Choi, Kim
  and Kim}]{kwak2011thyroid}
\bibinfo{author}{Kwak, J.Y.}, \bibinfo{author}{Han, K.H.},
  \bibinfo{author}{Yoon, J.H.}, \bibinfo{author}{Moon, H.J.},
  \bibinfo{author}{Son, E.J.}, \bibinfo{author}{Park, S.H.},
  \bibinfo{author}{Jung, H.K.}, \bibinfo{author}{Choi, J.S.},
  \bibinfo{author}{Kim, B.M.}, \bibinfo{author}{Kim, E.K.},
  \bibinfo{year}{2011}.
\newblock \bibinfo{title}{Thyroid imaging reporting and data system for {US}
  features of nodules: a step in establishing better stratification of cancer
  risk}.
\newblock \bibinfo{journal}{Radiology} \bibinfo{volume}{260},
  \bibinfo{pages}{892--899}.
%Type = Article
\bibitem[{Liu et~al.(2024)Liu, Zhong, Lin, Zhao, Fu and Liu}]{liu2024feature}
\bibinfo{author}{Liu, D.}, \bibinfo{author}{Zhong, S.}, \bibinfo{author}{Lin,
  L.}, \bibinfo{author}{Zhao, M.}, \bibinfo{author}{Fu, X.},
  \bibinfo{author}{Liu, X.}, \bibinfo{year}{2024}.
\newblock \bibinfo{title}{Feature-level {SMOTE}: Augmenting fault samples in
  learnable feature space for imbalanced fault diagnosis of gas turbines}.
\newblock \bibinfo{journal}{Expert Systems with Applications}
  \bibinfo{volume}{238}, \bibinfo{pages}{122023}.
%Type = Article
\bibitem[{Marcot and Hanea(2021)}]{marcot2021optimal}
\bibinfo{author}{Marcot, B.G.}, \bibinfo{author}{Hanea, A.M.},
  \bibinfo{year}{2021}.
\newblock \bibinfo{title}{What is an optimal value of k in k-fold
  cross-validation in discrete bayesian network analysis?}
\newblock \bibinfo{journal}{Computational Statistics} \bibinfo{volume}{36},
  \bibinfo{pages}{2009--2031}.
%Type = Article
\bibitem[{Mazo et~al.(2017)Mazo, Alegre and Trujillo}]{MAZO20171}
\bibinfo{author}{Mazo, C.}, \bibinfo{author}{Alegre, E.},
  \bibinfo{author}{Trujillo, M.}, \bibinfo{year}{2017}.
\newblock \bibinfo{title}{Classification of cardiovascular tissues using {LBP}
  based descriptors and a cascade {SVM}}.
\newblock \bibinfo{journal}{Computer Methods and Programs in Biomedicine}
  \bibinfo{volume}{147}, \bibinfo{pages}{1--10}.
%Type = Article
\bibitem[{Mohanty et~al.(2019)Mohanty, Sanket and Mishra}]{mohanty2019role}
\bibinfo{author}{Mohanty, J.}, \bibinfo{author}{Sanket},
  \bibinfo{author}{Mishra, P.}, \bibinfo{year}{2019}.
\newblock \bibinfo{title}{Role of {ACR-TIRADS} in risk stratification of
  thyroid nodules}.
\newblock \bibinfo{journal}{International Journal of Research in Medical
  Sciences} \bibinfo{volume}{7}, \bibinfo{pages}{1039--1043}.
%Type = Article
\bibitem[{Morin et~al.(2015)Morin, Basarab, Bidon and
  Kouam{\'e}}]{morin2015motion}
\bibinfo{author}{Morin, R.}, \bibinfo{author}{Basarab, A.},
  \bibinfo{author}{Bidon, S.}, \bibinfo{author}{Kouam{\'e}, D.},
  \bibinfo{year}{2015}.
\newblock \bibinfo{title}{Motion estimation-based image enhancement in
  ultrasound imaging}.
\newblock \bibinfo{journal}{Ultrasonics} \bibinfo{volume}{60},
  \bibinfo{pages}{19--26}.
%Type = Article
\bibitem[{Nguyen et~al.(2020)Nguyen, Kang, Pham, Batchuluun and
  Park}]{nguyen2020ultrasound}
\bibinfo{author}{Nguyen, D.T.}, \bibinfo{author}{Kang, J.K.},
  \bibinfo{author}{Pham, T.D.}, \bibinfo{author}{Batchuluun, G.},
  \bibinfo{author}{Park, K.R.}, \bibinfo{year}{2020}.
\newblock \bibinfo{title}{Ultrasound image-based diagnosis of malignant thyroid
  nodule using artificial intelligence}.
\newblock \bibinfo{journal}{Sensors} \bibinfo{volume}{20},
  \bibinfo{pages}{1822}.
%Type = Article
\bibitem[{Nguyen et~al.(2019)Nguyen, Pham, Batchuluun, Yoon and
  Park}]{nguyen2019artificial}
\bibinfo{author}{Nguyen, D.T.}, \bibinfo{author}{Pham, T.D.},
  \bibinfo{author}{Batchuluun, G.}, \bibinfo{author}{Yoon, H.S.},
  \bibinfo{author}{Park, K.R.}, \bibinfo{year}{2019}.
\newblock \bibinfo{title}{Artificial intelligence-based thyroid nodule
  classification using information from spatial and frequency domains}.
\newblock \bibinfo{journal}{Journal of Clinical Medicine} \bibinfo{volume}{8},
  \bibinfo{pages}{1976}.
%Type = Article
\bibitem[{Otsu(1979)}]{otsu1979threshold}
\bibinfo{author}{Otsu, N.}, \bibinfo{year}{1979}.
\newblock \bibinfo{title}{A threshold selection method from gray-level
  histograms}.
\newblock \bibinfo{journal}{IEEE Transactions on Systems, Man, and Cybernetics}
  \bibinfo{volume}{9}, \bibinfo{pages}{62--66}.
%Type = Article
\bibitem[{Panetta et~al.(2011)Panetta, Zhou, Agaian and
  Jia}]{panetta2011nonlinear}
\bibinfo{author}{Panetta, K.}, \bibinfo{author}{Zhou, Y.},
  \bibinfo{author}{Agaian, S.}, \bibinfo{author}{Jia, H.},
  \bibinfo{year}{2011}.
\newblock \bibinfo{title}{Nonlinear unsharp masking for mammogram enhancement}.
\newblock \bibinfo{journal}{IEEE Transactions on Information Technology in
  Biomedicine} \bibinfo{volume}{15}, \bibinfo{pages}{918--928}.
%Type = Inproceedings
\bibitem[{Pedraza et~al.(2015)Pedraza, Vargas, Narv{\'a}ez, Dur{\'a}n,
  Mu{\~n}oz and Romero}]{pedraza2015open}
\bibinfo{author}{Pedraza, L.}, \bibinfo{author}{Vargas, C.},
  \bibinfo{author}{Narv{\'a}ez, F.}, \bibinfo{author}{Dur{\'a}n, O.},
  \bibinfo{author}{Mu{\~n}oz, E.}, \bibinfo{author}{Romero, E.},
  \bibinfo{year}{2015}.
\newblock \bibinfo{title}{An open access thyroid ultrasound image database},
  in: \bibinfo{booktitle}{10th International Symposium on Medical Information
  Processing and Analysis}, \bibinfo{organization}{SPIE}. pp.
  \bibinfo{pages}{188--193}.
%Type = Article
\bibitem[{Poudel et~al.(2019)Poudel, Illanes, Ataide, Esmaeili, Balakrishnan
  and Friebe}]{poudel2019thyroid}
\bibinfo{author}{Poudel, P.}, \bibinfo{author}{Illanes, A.},
  \bibinfo{author}{Ataide, E.J.}, \bibinfo{author}{Esmaeili, N.},
  \bibinfo{author}{Balakrishnan, S.}, \bibinfo{author}{Friebe, M.},
  \bibinfo{year}{2019}.
\newblock \bibinfo{title}{Thyroid ultrasound texture classification using
  autoregressive features in conjunction with machine learning approaches}.
\newblock \bibinfo{journal}{IEEE Access} \bibinfo{volume}{7},
  \bibinfo{pages}{79354--79365}.
%Type = Article
\bibitem[{Raj et~al.(2025)Raj, Ahuja and Busnel}]{RAJ2025104910}
\bibinfo{author}{Raj, A.}, \bibinfo{author}{Ahuja, K.},
  \bibinfo{author}{Busnel, Y.}, \bibinfo{year}{2025}.
\newblock \bibinfo{title}{{AI} algorithm for predicting and optimizing
  trajectory of massive {UAV} swarm}.
\newblock \bibinfo{journal}{Robotics and Autonomous Systems}
  \bibinfo{volume}{186}, \bibinfo{pages}{104910}.
%Type = Article
\bibitem[{Rastogi et~al.(2024)Rastogi, Johri, Tiwari and
  Elngar}]{rastogi2024multi}
\bibinfo{author}{Rastogi, D.}, \bibinfo{author}{Johri, P.},
  \bibinfo{author}{Tiwari, V.}, \bibinfo{author}{Elngar, A.A.},
  \bibinfo{year}{2024}.
\newblock \bibinfo{title}{Multi-class classification of brain tumour magnetic
  resonance images using multi-branch network with inception block and
  five-fold cross validation deep learning framework}.
\newblock \bibinfo{journal}{Biomedical Signal Processing and Control}
  \bibinfo{volume}{88}, \bibinfo{pages}{105602}.
%Type = Article
\bibitem[{Saini et~al.(2024)Saini, Ahuja and Chauhan}]{saini2024improved}
\bibinfo{author}{Saini, S.}, \bibinfo{author}{Ahuja, K.},
  \bibinfo{author}{Chauhan, A.S.}, \bibinfo{year}{2024}.
\newblock \bibinfo{title}{Improved and explainable cervical cancer
  classification using ensemble pooling of block fused descriptors}.
\newblock \bibinfo{journal}{arXiv preprint arXiv:2405.01600} .
%Type = Article
\bibitem[{Shastri et~al.(2018)Shastri, Tamrakar and Ahuja}]{shastri2018density}
\bibinfo{author}{Shastri, A.A.}, \bibinfo{author}{Tamrakar, D.},
  \bibinfo{author}{Ahuja, K.}, \bibinfo{year}{2018}.
\newblock \bibinfo{title}{Density-wise two stage mammogram classification using
  texture exploiting descriptors}.
\newblock \bibinfo{journal}{Expert Systems with Applications}
  \bibinfo{volume}{99}, \bibinfo{pages}{71--82}.
%Type = Article
\bibitem[{Singh et~al.(2007)Singh, Kumar and Verma}]{singh2007optimization}
\bibinfo{author}{Singh, S.}, \bibinfo{author}{Kumar, V.},
  \bibinfo{author}{Verma, H.}, \bibinfo{year}{2007}.
\newblock \bibinfo{title}{Optimization of block size for {DCT-based} medical
  image compression}.
\newblock \bibinfo{journal}{Journal of Medical Engineering \& Technology}
  \bibinfo{volume}{31}, \bibinfo{pages}{129--143}.
%Type = Article
\bibitem[{Song et~al.(2018)Song, Li, Liu, Qin, Zhang, Zhang and
  Hao}]{song2018multitask}
\bibinfo{author}{Song, W.}, \bibinfo{author}{Li, S.}, \bibinfo{author}{Liu,
  J.}, \bibinfo{author}{Qin, H.}, \bibinfo{author}{Zhang, B.},
  \bibinfo{author}{Zhang, S.}, \bibinfo{author}{Hao, A.}, \bibinfo{year}{2018}.
\newblock \bibinfo{title}{Multitask cascade convolution neural networks for
  automatic thyroid nodule detection and recognition}.
\newblock \bibinfo{journal}{IEEE Journal of Biomedical and Health Informatics}
  \bibinfo{volume}{23}, \bibinfo{pages}{1215--1224}.
%Type = Article
\bibitem[{Srivastava and Kumar(2022)}]{srivastava2022hybrid}
\bibinfo{author}{Srivastava, R.}, \bibinfo{author}{Kumar, P.},
  \bibinfo{year}{2022}.
\newblock \bibinfo{title}{A hybrid model for the identification and
  classification of thyroid nodules in medical ultrasound images}.
\newblock \bibinfo{journal}{International Journal of Modelling, Identification
  and Control} \bibinfo{volume}{41}, \bibinfo{pages}{32--42}.
%Type = Article
\bibitem[{Srivastava and Kumar(2024)}]{srivastava2024deep}
\bibinfo{author}{Srivastava, R.}, \bibinfo{author}{Kumar, P.},
  \bibinfo{year}{2024}.
\newblock \bibinfo{title}{{Deep-GAN}: an improved model for thyroid nodule
  identification and classification}.
\newblock \bibinfo{journal}{Neural Computing and Applications}
  \bibinfo{volume}{36}, \bibinfo{pages}{7685--7704}.
%Type = Article
\bibitem[{Tasnimi and Ghaffari(2023)}]{tasnimi2023diagnosis}
\bibinfo{author}{Tasnimi, M.}, \bibinfo{author}{Ghaffari, H.R.},
  \bibinfo{year}{2023}.
\newblock \bibinfo{title}{Diagnosis of anomalies based on hybrid features
  extraction in thyroid images}.
\newblock \bibinfo{journal}{Multimedia Tools and Applications}
  \bibinfo{volume}{82}, \bibinfo{pages}{3859--3877}.
%Type = Inproceedings
\bibitem[{Ullah et~al.(2020)Ullah, Gupta, Ahuja, Tiwari and Kumar}]{9116373}
\bibinfo{author}{Ullah, S.}, \bibinfo{author}{Gupta, S.},
  \bibinfo{author}{Ahuja, K.}, \bibinfo{author}{Tiwari, A.},
  \bibinfo{author}{Kumar, A.}, \bibinfo{year}{2020}.
\newblock \bibinfo{title}{{L2L}: A highly accurate {Log\_2\_Lead} quantization
  of pre-trained neural networks}, in: \bibinfo{booktitle}{2020 Design,
  Automation \& Test in Europe Conference \& Exhibition (DATE)}, pp.
  \bibinfo{pages}{979--982}.
%Type = Article
\bibitem[{Wang et~al.(2025)Wang, Feng, Wang and Niu}]{wang2025robust}
\bibinfo{author}{Wang, X.}, \bibinfo{author}{Feng, L.}, \bibinfo{author}{Wang,
  D.}, \bibinfo{author}{Niu, P.}, \bibinfo{year}{2025}.
\newblock \bibinfo{title}{A robust wavelet domain multi-scale texture
  descriptor for image classification}.
\newblock \bibinfo{journal}{Expert Systems with Applications}
  \bibinfo{volume}{265}, \bibinfo{pages}{126000}.
%Type = Inproceedings
\bibitem[{Wu and Liu(2019)}]{wu2019classification}
\bibinfo{author}{Wu, Y.}, \bibinfo{author}{Liu, P.}, \bibinfo{year}{2019}.
\newblock \bibinfo{title}{A classification algorithm of ultrasonic thyroid
  standard planes using {LBP} and {HOG} features}, in: \bibinfo{booktitle}{2019
  IEEE 13th International Conference on Anti-counterfeiting, Security, and
  Identification (ASID)}, \bibinfo{organization}{IEEE}. pp.
  \bibinfo{pages}{103--107}.
%Type = Article
\bibitem[{Zouhri et~al.(2022)Zouhri, Homri and Dantan}]{zouhri2022handling}
\bibinfo{author}{Zouhri, W.}, \bibinfo{author}{Homri, L.},
  \bibinfo{author}{Dantan, J.Y.}, \bibinfo{year}{2022}.
\newblock \bibinfo{title}{Handling the impact of feature uncertainties on
  {SVM}: A robust approach based on {Sobol} sensitivity analysis}.
\newblock \bibinfo{journal}{Expert Systems with Applications}
  \bibinfo{volume}{189}, \bibinfo{pages}{115691}.

\end{thebibliography}

\end{document}